\pdfoutput=1
\documentclass[11pt]{article}

\usepackage[final]{acl}

\usepackage{blindtext}
\usepackage{hyperref}
\usepackage{enumitem}

\usepackage{times}
\usepackage{latexsym}
\usepackage{subcaption}

\usepackage[T1]{fontenc}
\usepackage[utf8]{inputenc}

\usepackage{microtype}

\usepackage{inconsolata}

\usepackage{graphicx}
\usepackage{booktabs}
\usepackage{mathtools} 
\usepackage{amsmath}
\usepackage{xcolor}
\usepackage{enumitem}

\title{RoSE: Round-robin Synthetic Data Evaluation for Selecting LLM Generators without Human Test Sets}

\author{Jan Cegin$^\dagger$, Branislav Pecher$^\dagger$, Ivan Srba$^\dagger$, Jakub Simko$^\dagger$ \\
  $^\dagger$ Kempelen Institute of Intelligent Technologies, Bratislava, Slovakia\\
  \texttt{\{jan.cegin, branislav.pecher, jakub.simko, ivan.srba\}}@kinit.sk \\}

\begin{document}
\maketitle
\begin{abstract}
LLMs are powerful generators of synthetic data, which are used for training smaller, specific models. This is especially valuable for low-resource languages, where human-labelled data is scarce but LLMs can still produce high-quality text. However, LLMs differ in how useful their outputs are for training. Selecting the best LLM as a generator is challenging because extrinsic evaluation requires costly human annotations (which are often unavailable for low-resource languages), while intrinsic metrics correlate poorly with downstream performance. We introduce Round-robin Synthetic data Evaluation (RoSE), a proxy metric for selecting the best LLM generator without human test sets. RoSE trains a small model on the outputs of a candidate generator (LLM) and then evaluates it on generated synthetic examples from all other candidate LLMs. The final RoSE score is the mean performance of this small model. Across six LLMs, eleven languages, and three tasks (sentiment, topic, intent), RoSE identifies the optimal generator more often than any other intrinsic heuristics. RoSE outperforms intrinsic heuristics and comes within 0.76 percentage points of the optimal generator baseline. This result is measured in terms of downstream performance, obtained by training a small model on the chosen generator’s outputs (optimal vs. proxy-metric–selected) and evaluating it on human-labelled test data. Additionally, RoSE is the only metric to achieve a positive correlation with performance on human test data.
  
\end{abstract}

\maketitle

\section{Introduction}
\begin{figure}[t!]
    \centering
    \includegraphics[width=6.5cm]{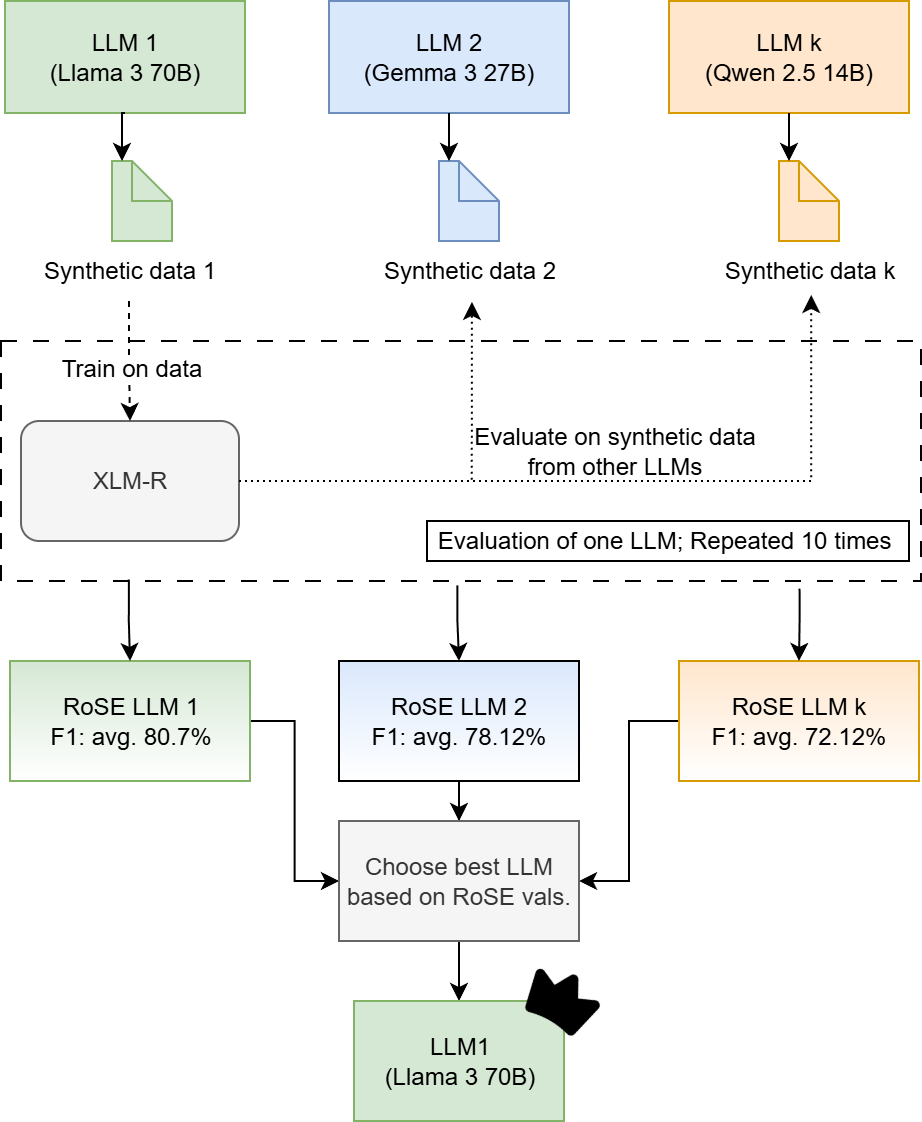}
    \caption{Overview of RoSE proxy metric calculations. For each candidate generator (LLM), we first generate synthetic data. Then, we train a smaller model on each generator’s synthetic dataset and evaluate it on the synthetic data generated by all other generators. The mean F1 score across these test sets is the final RoSE score for that LLM. The highest RoSE score LLM is considered the best generator.}
    \label{fig:methodology}
\end{figure}

Current large language models (LLMs), such as GPT4, Llama, and others, show impressive performance in generating well-formed texts~\citep{cegin-etal-2023-chatgpt}. Such synthetic texts are commonly leveraged to train smaller, more efficient downstream models, enhancing their performance~\citep{wang-etal-2025-diversity, cegin2024llmsvsestablishedtext}. While most of the previous research has focused on improving the LLMs' ability to generate representative synthetic data in English~\citep{piedboeuf-langlais-2023-chatgpt, cegin2024effectsdiversityincentivessample, wang-etal-2025-diversity}, some research has also focused on generative strategies and LLM evaluation for low-resource languages~\citep{anikina2025rigorousevaluationllmdata}. Even for low-resource languages, LLMs can produce high-quality samples~\citep{son2025efficient, chung2025beyond, anikina2025rigorousevaluationllmdata}. However, previous research has identified that differences exist between LLMs when generating texts in low-resource languages, as their quality for training smaller models can vary significantly~\citep{anikina2025rigorousevaluationllmdata}. It is therefore important to have measures for evaluating the LLM's performance for generating textual data.

Evaluation methods for the purpose of identifying the best LLM generator generally fall into two categories: intrinsic and extrinsic metrics~\citep{li-etal-2024-llatrieval}. Intrinsic evaluation measures properties of the generated text itself, such as vocabulary size, type–token ratio, or diversity, without training a downstream model (and without much dependence on the task). Extrinsic evaluation, in contrast, assesses the usefulness of synthetic data indirectly, by training a downstream model on it and evaluating performance on human-labelled test data. As this relies on human annotations, extrinsic evaluation is treated as an oracle, the gold standard for measuring the true utility of generated data~\citep{cegin-etal-2023-chatgpt, li-etal-2024-llatrieval}, though it is rarely available in low-resource settings~\cite{anikina2025rigorousevaluationllmdata}.

The challenge is that for many low-resource language-task combinations, human test sets do not exist~\cite{anikina2025rigorousevaluationllmdata}, or, at most, only a very small number of human samples ($\approx$10 per label) exist. In these settings, evaluation of LLMs as synthetic data generators and their selection relies only on intrinsic metrics or simple heuristics. Yet prior work shows that intrinsic metrics do not reliably correlate with extrinsic outcomes even for English data: correlations may be weak~\citep{wang-etal-2025-diversity} or inconsistent~\citep{cegin2024llmsvsestablishedtext}. As the quality of generated data in low-resource languages can vary widely across various LLMs used as generators~\citep{anikina2025rigorousevaluationllmdata}, \textbf{reliable proxy metrics are needed to identify the best LLM for synthetic data generation when human evaluation data is unavailable}.

For such cases, when human evaluation data is unavailable, we propose our proxy metric, Round-robin Synthetic data Evaluation (RoSE), visualised in Figure~\ref{fig:methodology}, for evaluating LLMs as generators and identifying the best generator, where human test data is not available. The intuition is that synthetic data carries the representational signature of its generator, and since different LLMs produce data of varying quality and coverage~\cite{anikina2025rigorousevaluationllmdata}, cross-evaluating on each other’s outputs could reveal which generator provides the most generalizable training signal. First, given a set of evaluated LLMs as generators from which we aim to find the best generator, we generate synthetic text for each task–language combination per each LLM via the current state-of-the-art method, which leverages a small (10 per label) amount of human examples. Second, a smaller model is trained on one generator’s data and evaluated on synthetic test sets from all other generators. The performance of that LLM as a generator is represented as the mean over all the synthetic test sets. This process is repeated for all candidate LLMs 10 times, where the best LLM is chosen based on its highest mean performance.

To evaluate RoSE, we conduct a comparative analysis of proxy metrics, including intrinsic metrics, simple heuristics, and our proposed approach (RoSE). Experiments span 11 typologically diverse languages with varied scripts, covering several very low-resource cases such as Welsh, Romanian, Azerbaijani, etc., across three classification tasks (sentiment, topic, and intent). We evaluate six open-weight LLMs of different sizes and families on each task–language combination. To test whether a proxy metric can replace human-labelled data for identifying the best generator LLM, we treat the generator selected using human validation sets as an optimal generator. For each proxy metric, we measure the F1 performance gap between (a) a small model trained on data from the proxy-selected generator and (b) a model trained on data from the optimally-selected generator and evaluated on human test data. We also report Pearson correlations between each metric’s scores and the corresponding LLMs' human test performance, as well as ranking-based correlations.

Our findings show that:
\begin{itemize}
    \item Across all languages and tasks, RoSE identifies the optimal LLM generator more often than any other proxy metric.
    \item LLM generators selected by RoSE achieve an average gap of only 0.76\% F1 compared to the optimal human-performance-based generator selection, versus 2.52\% for the second-best proxy metric.  
    \item RoSE is the only proxy metric that consistently yields a positive correlation between classifier performance and human evaluation.  
    \item RoSE ranks best across 9 of 11 languages, and second-best in the remaining 2.  
\end{itemize}

Additional ablations further demonstrate that RoSE remains effective even when comparing as few as three LLMs. RoSE also performs strongly when comparing LLMs of similar parameter size. We also further analyse how the number of LLMs used for computing RoSE affects selection quality and how the generation setup influences RoSE’s performance. We release all our code, data and results in \url{https://github.com/kinit-sk/RoSE}. %\textit{see attached ZIP file}.

\section{Related Work}
Since their introduction, large language models (LLMs) such as GPT-4 and Llama have been increasingly adopted as tools for data augmentation and generation. The synthetic data they produce is commonly used to train smaller downstream models, improving efficiency while maintaining strong performance. This approach has been applied across a wide range of tasks, including automated scoring~\cite{fang2023using}, intent classification~\cite{sahu-etal-2022-data}, sentiment analysis~\cite{piedboeuf-langlais-2023-chatgpt, ONAN2023101611, yoo-etal-2021-gpt3mix-leveraging}, hate speech detection~\cite{sen-etal-2023-people}, news classification~\cite{piedboeuf-langlais-2023-chatgpt}, content recommendation~\cite{contect-based-recom}, and health symptom classification~\cite{dai2023auggpt}. 

While most of the generation has been done in English, recent research has also focused on the usage of LLM for generating synthetic texts in low-resource languages. Multilingual synthetic generation using LLMs has been leveraged for various tasks like QA~\cite{kramchaninova-defauw-2022-synthetic, namboori2023gemquad, putri-etal-2024-llm}, fact-checking~\cite{chung2025beyond}, reasoning~\cite{pranida2025syntheticdatagenerationculturally}, NER~\cite{liu-etal-2021-mulda}, sentiment stance detection~\cite{ZOTOVA2021114547} and classification~\cite{glenn-etal-2023-jetsons, anikina2025rigorousevaluationllmdata}. 

Recent work has introduced efforts to benchmark LLMs as synthetic data generators, most notably with AgoraBench~\cite{kim-etal-2025-evaluating}. While this benchmark is effective at identifying top-performing LLMs for generation tasks, its primary focus lies in post-training decoder-based models on synthetic data rather than training downstream encoder-based models. Currently, there is no established benchmarking framework for selecting the most suitable LLM generator for downstream classification tasks. Such selection is often performed on a case-by-case basis using extrinsic evaluation on human-annotated test sets corresponding to the target task~\cite{glenn-etal-2023-jetsons, anikina2025rigorousevaluationllmdata}. However, human-labelled test sets may be unavailable for many tasks and languages, and their creation can be prohibitively expensive~\cite{putri-etal-2024-llm, pranida2025syntheticdatagenerationculturally, gurgurov-etal-2025-gremlin}. Thus, identifying optimal LLM generators through reliable proxy metrics becomes crucial, as it avoids the high cost of human evaluation.

\section{Methodology and Experiments}

\subsection{Round-robin Synthetic Data Evaluation (RoSE)}

We propose \textbf{Round-robin Synthetic data Evaluation (RoSE)} as a proxy metric for identifying the best large language model (LLM) for synthetic data generation in the absence of human test sets. The metric calculations are visualised in Figure~\ref{fig:methodology}.

Synthetic data produced by LLMs carries the representational signature of its generator, with different models exhibiting substantial variation in coverage and quality when generating texts in low-resource languages~\citep{anikina2025rigorousevaluationllmdata}. A robust generator should produce data that transfers well across distributions, such that a classifier trained on its outputs generalises effectively to data generated by other LLMs. This intuition motivates RoSE as a proxy for extrinsic evaluation, where human-labelled test sets act as an oracle.

To compute the method, we start with a given set of candidate LLMs, and we proceed in three steps in a round-robin manner. First, for each task--language combination, every LLM generates synthetic training and test data using a state-of-the-art generation procedure. Second, a small classifier is trained on the synthetic data produced by one LLM and evaluated on the synthetic test sets generated by all other LLMs. Third, the performance of the LLM is defined as the mean score of this classifier across all cross-evaluations. The process is repeated for all candidate LLMs 10 times (to mitigate randomness as per~\cite{pecher2023effects}), and the generator achieving the highest mean performance is selected as the best model.

\subsection{Intrinsic Metrics}\label{sec:intrinsic_metrics}
We use a wide variety of 8 different intrinsic metrics to evaluate their effectiveness against our proposed RoSE metric. The tokenisation is done via the \textit{XLMRoberta-base} tokeniser. We did not use the MAUVE~\citep{pillutla2021mauve} metric which measures the gap between neural text and human text, as the human distribution we used was very small (10 samples per label) and early testing showed that MAUVE did not produce results better than random chance in such setting.

\textbf{Average Pairwise Cosine Distance:} Embeddings are computed for all samples within each label and measure the average cosine distance between sample pairs. This captures the semantic diversity~\citep{reimers-gurevych-2019-sentence} of generations conditioned on the same label\footnote{The model used for this is: \textit{paraphrase-multilingual-MiniLM-L12-v2}}.

\textbf{Bigram Diversity:} This measures~\citep{li-etal-2016-diversity} the proportion of distinct bigram tokens relative to the total number of bigrams in the dataset, reflecting diversity at the phrase level.

\textbf{Number of Valid Samples:} We count the proportion of syntactically or semantically valid generations that remain after applying a revision step (as described in Section~\ref{sec:experiments}), reflecting robustness and generation quality.

\textbf{Silhouette Score:} Using embeddings, we compute the silhouette coefficient to assess clustering quality of the generated data, measuring how well samples group by their intended label compared to other labels~\citep{rousseeuw1987silhouettes}.

\textbf{Type–Token Ratio (TTR):} This metric~\citep{johnson1944type} quantifies lexical diversity by normalising the number of unique tokens with respect to the total number of tokens.

\textbf{Token Entropy:} We measure the entropy of the token distribution in the generated data, where higher entropy indicates greater lexical variety and unpredictability~\citep{rosillo2024entropy}

\textbf{LLM Parameter Size:} A simple heuristic that always chooses the LLM with the largest number of parameters, motivated by the observation that larger models often perform better.

\textbf{Random:} As a naive baseline, we select the best LLM uniformly at random across ten runs and average the resulting scores.

\subsection{Data and Tasks}\label{sec:data_tasks}

We evaluate RoSE and other proxy metrics on three classification tasks: intent recognition, topic classification, and sentiment analysis. For each task, we generate synthetic data in 11 typologically diverse languages. The selection includes two high-resource languages (English, German), four mid-resource languages (Thai, Hebrew, Indonesian, Swahili), and five low-resource languages (Romanian, Azerbaijani, Slovenian, Telugu, Welsh). This choice reflects both linguistic diversity and the availability of datasets for the target tasks.

For intent recognition, we use the MASSIVE dataset \cite{fitzgerald-etal-2023-massive}, a multilingual benchmark for virtual assistant evaluation covering 51 languages and 60 intents. To simplify the task, we restrict the label set to the ten most common intents.

For topic classification, we rely on SIB-200 \cite{adelani-etal-2024-sib}, which provides seven topic labels derived from the FLORES-200 machine translation corpus, annotated at the sentence level.

For sentiment classification, no single multilingual dataset spanning both high- and low-resource languages currently exists. We therefore combine ten datasets for low-resource languages from \cite{gurgurov-etal-2025-gremlin,gurgurov-etal-2024-adapting} with two additional datasets for English and German from \cite{mollanorozy-etal-2023-cross}. As noted in \cite{anikina2025rigorousevaluationllmdata}, these datasets vary in coverage and text domain: for instance, the German set focuses on transportation and infrastructure, while Romanian data primarily consists of product reviews.

\subsection{Models}\label{sec:models}

We use up to 6 LLMs of different sizes: Gemma-3~\cite{gemma_2025} with 4 and 27 billion parameters, Magistral \textit{Small} with 24 billion parameters, Qwen 2.5~\citep{qwen2.5} with 14 billion parameters, and Llama-3~\cite{llama3modelcard} with 8 and 70 billion parameters. These models were chosen for their open-weight nature and support for multiple languages. For finetuning a smaller model, we used the XLM-R \textit{Base}~\cite{DBLP:journals/corr/abs-1911-02116} model. 

\subsection{Experimental Setup}\label{sec:experiments}

We evaluate our proxy metric, RoSE, alongside the selected intrinsic metrics from Section~\ref{sec:intrinsic_metrics} across all task–language combinations (33 in total). For each combination, we first generate 100 samples per label from each LLM under comparison. To ensure strong baselines, we adopt the best-performing generation setup identified by~\citep{anikina2025rigorousevaluationllmdata}, which uses prompts with 10 in-context randomly selected human examples from the train set and applies self-revision as a filtering mechanism. We also test a different, worse-performing setup of excluding the in-context human examples. Results for this ablation can be found in Section~\ref{sec:no_icl_used}. Details about generation setup can be found in Appendix~\ref{sec:appendix_prompt_templates}.

Next, we fine-tune XLM-R~\citep{DBLP:journals/corr/abs-1911-02116} ten times for each task–language combination. The trained classifiers are then evaluated both on synthetic data generated by other LLMs (to compute RoSE) and on human-annotated data (to identify the best-performing LLM). Finally, after fine-tuning, we compute the remaining intrinsic metrics and report the corresponding F1 scores for each case. Details regarding the downstream model fine-tuning can be found in~\ref{sec:appendix_downstream_finetuning}.

To identify the best proxy metric for LLM selection, we proceed as follows. For each task–language case, we train classifiers on synthetic data generated by each LLM and evaluate them on human-annotated data. The resulting average F1 scores provide a ranking of the LLMs as generators, from worst to best, and enable us to identify the optimal LLM generator.
For each proxy metric, we then compute an alternative ranking of the LLMs based on their values. To assess the quality of these rankings, we measure three outcomes.
First, the correct identification: how many times was the optimal LLM generator identified by the proxy metric, and how many times were the best 3 LLM generators (irrespective of order) identified by the proxy metric. Second, the performance gap: the difference in F1 score between the downstream model trained with the LLM chosen by the proxy and the downstream model trained with the optimally-selected LLM (ideal outcome = 0\%).  
Third, the correlations: the Pearson correlation between the proxy metric values and the classifiers’ performance on human data. We also look at rank correlations via Kendal $\tau$ to compare order rankings for each proxy metric.

\section{Results and Discussion}

\begin{table}[t!]
\centering
\small
\begin{tabular}{lcc}
\toprule
Proxy Metric & Top-1 Match & Top-3 Match \\
\midrule
Avg. cos. dist. & 7 (21.21\%)  & 8 (24.24\%)  \\
Bigram Div. & 5 (15.15\%)  & 2 (6.06\%)  \\
No. Samples & 7 (21.21\%)  & 0 (0.00\%)  \\
Silhouette Score & 4 (12.12\%)  & 2 (6.06\%)  \\
Type Token Ratio & 3 (9.09\%)  & 4 (12.12\%)  \\
Token Entropy & 4 (12.12\%)  & 2 (6.06\%)  \\
LLM param. size & 12 (36.36\%) & 0 (0.00\%)  \\
Random & 5 (15.15\%) & 2 (6.06\%)  \\
RoSE (ours) & \textbf{20 (60.61\%)} & \textbf{15 (45.45\%)} \\
\bottomrule
\end{tabular}
\caption{Number of times each proxy metric correctly identified the optimal LLM generator (Top-1) or correctly identified the top 3 generators (Top-3, irrespective of order) across all tasks and languages. RoSE achieves the highest performance in both cases.}
\label{tab:metric_matches}
\end{table}

When comparing our RoSE metric with the proxy metrics using Top-1 and Top-3 evaluation in Table~\ref{tab:metric_matches}, Top-1 is achieved in 20 out of 33 cases (60.60\%), outperforming other proxy metrics (next best at 36.36\%). For Top-3, it matches the top 3 generators in 15 (45.45\%) cases compared to the next best proxy at 8 cases (24.24\%).

\begin{figure}[t!]
    \centering
    \includegraphics[width=7.7cm]{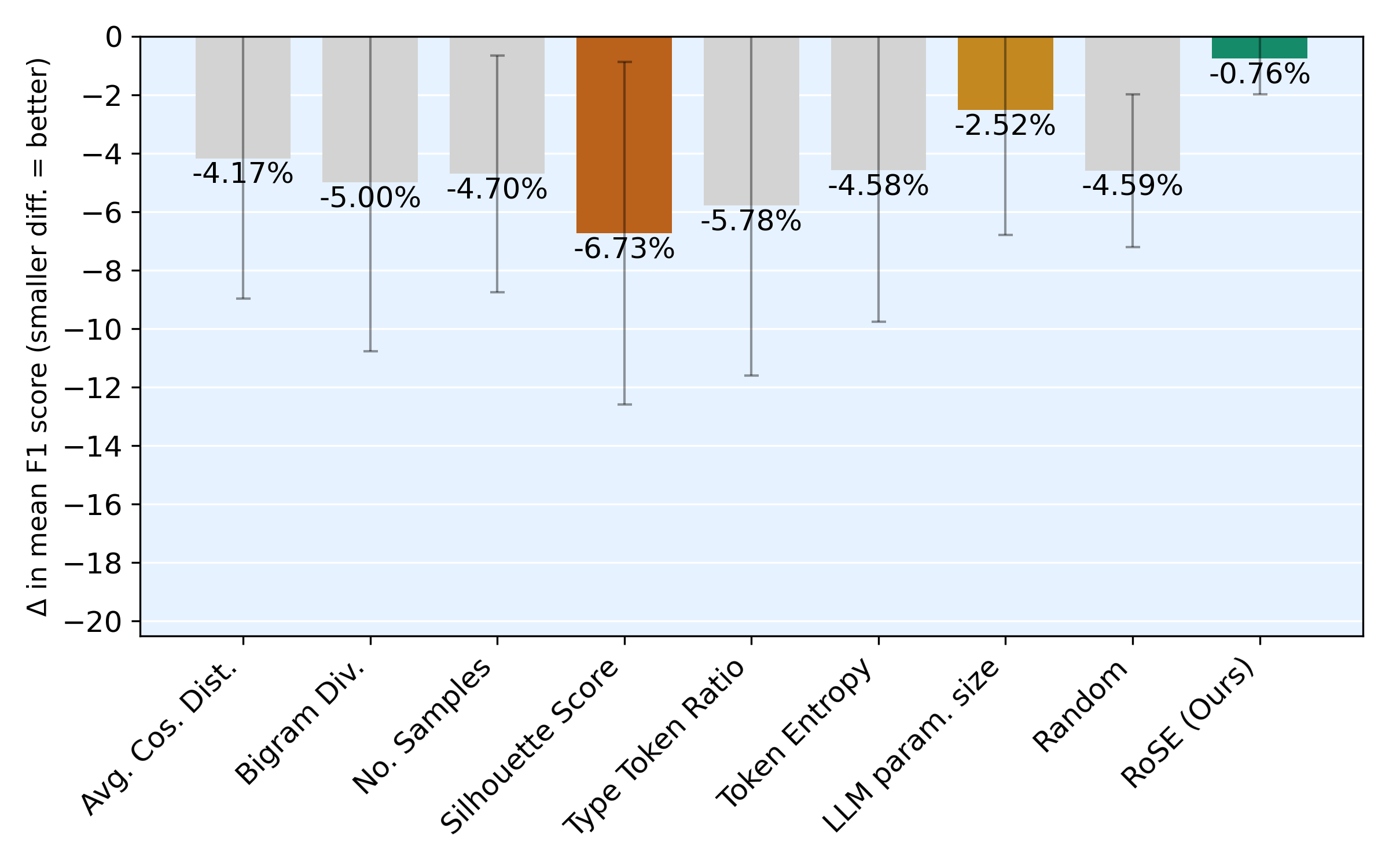}
    \caption{Comparison of proxy metrics for selecting the best LLM generator. Bars show the average gap in mean F1 score for models trained on the best generator selected by metrics vs. the optimal generator (smaller is better). The models are evaluated on human test data. The best metric is green, the second best is orange, and the worst is red.}
    \label{fig:results_main}
\end{figure}

Figure~\ref{fig:results_main} summarises the aggregated results across all tasks and languages when comparing six LLMs to identify the best generator. The reported values represent the performance gap in F1 score between downstream models trained on data from the LLM selected by a given metric and those trained on data from the optimal LLM generator.

In the cases where the optimal LLM is not selected, RoSE’s selected LLM yields a downstream model whose performance is very close to a model trained on data from the optimal generator, resulting in the lowest average performance gap among all evaluated metrics (0.76\%). We also note that on average, only 3 intrinsic metrics achieve statistically significantly (\textit{p=0.05} for Mann-Whitney U test) better generator selection than choosing a generator at random. Considering that the second-best metric is the LLM parameter size, which always selects the largest LLM as the generator based on its parameter size, we can also conclude that RoSE successfully identifies cases where the largest LLM is not the optimal generator, resulting in efficient solutions.

\begin{table}[t!]
\centering
\small
\begin{tabular}{lccc}
\toprule
Proxy Metric & Intent & Topic & Sentiment \\
\midrule
Avg. cos. dist.   & -2.84\% & -2.58\% & -7.10\% \\
Bigram Div.       & -3.05\% & -5.27\% & -6.67\% \\
No. Samples       & -3.26\% & -4.82\% & -6.02\% \\
Silhouette Score  & -2.70\% & -8.02\% & -9.47\% \\
Type Token Ratio  & -3.40\% & -5.59\% & -8.35\% \\
Token Entropy     & -3.48\% & -6.09\% & \underline{-4.17\%} \\
LLM param. size   & \underline{-1.73}\% & \textbf{-0.31\%} & -5.52\% \\
Random            & -3.05\% & -4.65\% & -6.07\% \\
RoSE (Ours)       & \textbf{-0.64\%} & \underline{-0.77\%} & \textbf{-0.86\%} \\
\bottomrule
\end{tabular}
\caption{Comparison of proxy metrics for selecting the best LLM generator per task. The table shows the gap in the mean F1 score for models trained on the best generator selected by metrics vs. the optimal generator (smaller is better). The best metric is bold, the second best is underlined. RoSE performs well across all tasks as either the best or the second-best proxy metric.}
\label{tab:proxy_delta}
\end{table}

\begin{table*}[t!]
\centering
\small
\begin{tabular}{lccccccccccc}
\toprule
Proxy Metric & az & cy & he & th & sw & sl & en & de & id & ro & te \\
\midrule
Type Token Ratio & -5.07 & -8.62 & -14.82 & -4.05 & -6.56 & -7.95 & -3.80 & \underline{-0.95} & -5.69 & -2.65 & \underline{-3.41} \\
Bigram Div. & -3.30 & -6.56 & -14.82 & \underline{-1.36} & -6.56 & -7.85 & -3.80 & -1.65 & -3.00 & -2.65 & \underline{-3.41} \\
Token Entropy & -5.68 & -6.56 & -14.82 & -2.39 & -6.56 & -2.43 & \underline{-1.22} & -1.65 & -3.00 & -2.65 & \underline{-3.41} \\
Silhouette Score & -4.71 & -12.76 & -5.60 & -9.83 & -4.84 & -3.55 & -7.57 & -4.95 & -4.09 & -3.52 & -12.62 \\
Avg. cos. dist. & -4.58 & -8.48 & -6.55 & \textbf{0.00} & -2.51 & -7.95 & -3.80 & -3.91 & \textbf{-1.09} & -1.86 & -5.17 \\
No. Samples & -5.41 & -2.44 & \underline{-1.13} & -6.47 & -4.98 & \underline{-0.71} & -11.51 & -5.39 & -3.82 & -6.11 & -3.73 \\
LLM param. size & \underline{-2.03} & \textbf{-0.77} & -2.45 & -2.91 & \underline{-2.33} & -1.34 & -2.22 & -4.31 & -1.14 & \underline{-0.78} & -7.43 \\
Random & -3.61 & -5.96 & -6.35 & -4.28 & -5.34 & -2.48 & -4.90 & -3.01 & -3.87 & -4.36 & -6.29 \\
RoSE (ours) & \textbf{-1.77 }& \underline{-0.90} & \textbf{0.00} & \textbf{0.00} & \textbf{-1.82} & \textbf{-0.43} & \textbf{-0.68} & \textbf{0.00} & \underline{-1.11} & \textbf{-0.43} & \textbf{-1.19} \\
\bottomrule
\end{tabular}
\caption{Mean gap between F1 performance for downstream models trained on the LLM generator selected by proxy metric vs. optimal LLM generator. The lower the gap, the better the proxy metric. The best proxy for language is bolded, and the second-best is underlined. The best proxy metric for each language is bolded.}
\label{tab:metric_means}
\end{table*}

\begin{figure*}[t!]
\centering
\begin{subfigure}{0.49\textwidth}
\centering
\includegraphics[width = \textwidth]{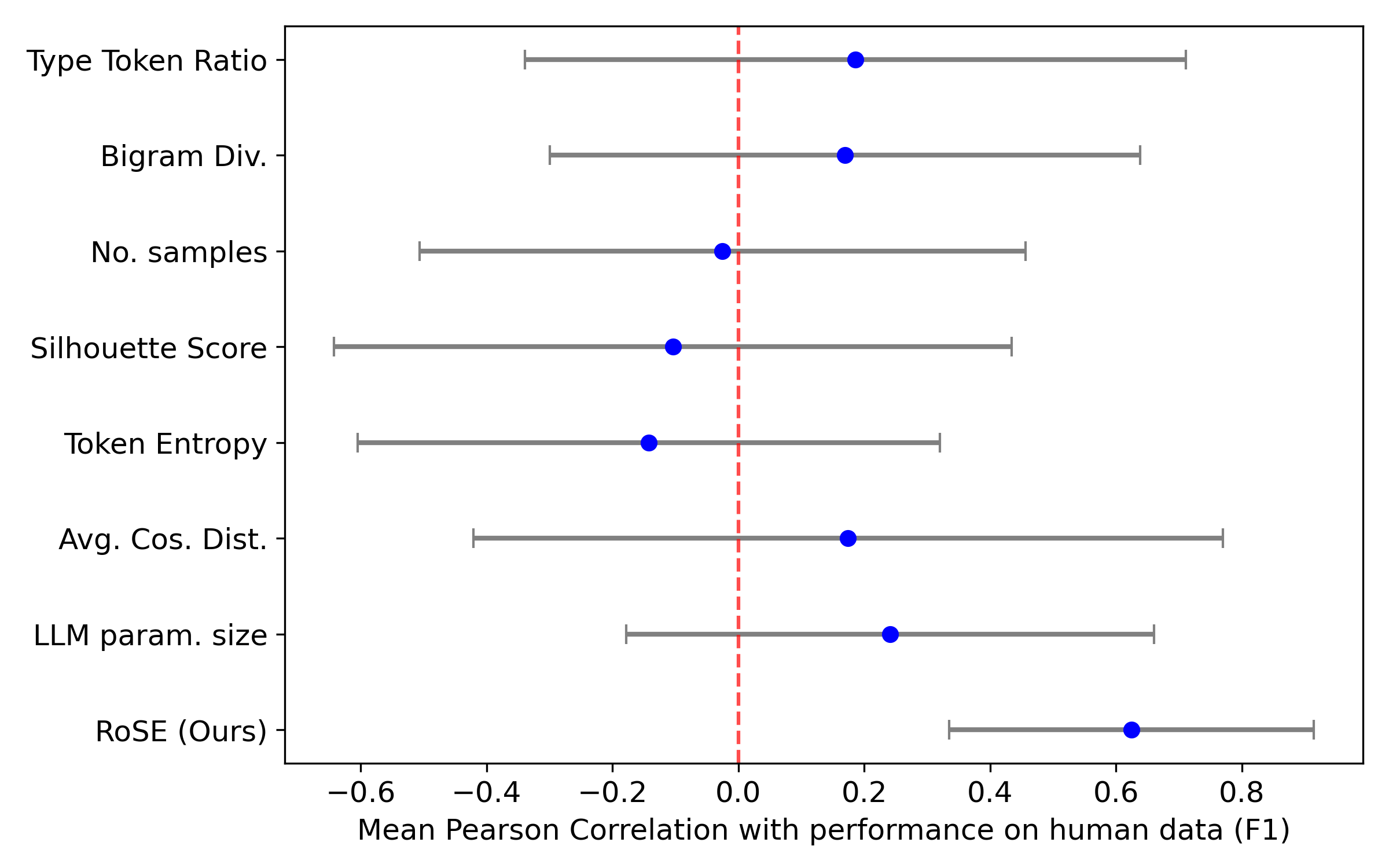}
\caption{Pearson correlation.}
\label{fig:results_corrs}
\end{subfigure}
\begin{subfigure}{0.49\textwidth}
\centering
\includegraphics[width = \textwidth]{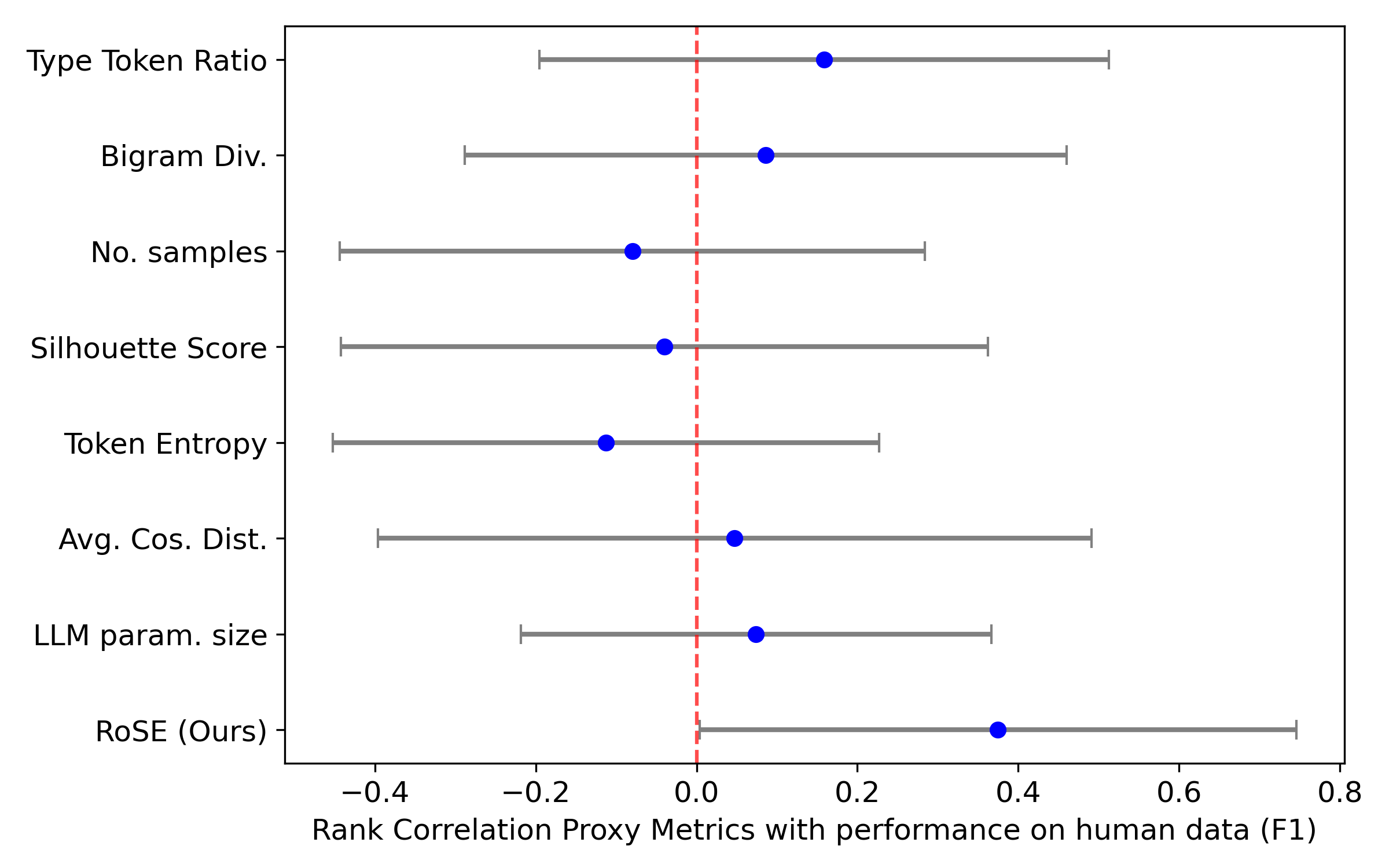}
\caption{Rank (Kendal $\tau$) correlation.}
\label{fig:results_corrs_kendal}
\end{subfigure}
\caption{Forest plot showing mean Pearson and Rank (Kendal $\tau$) correlations between proxy metrics (except for random, as it has no values) and downstream performance of models on human-labelled data (F1). Error bars denote confidence intervals. RoSE is the most reliable proxy metric; its average Pearson and rank correlations with human F1 are clearly higher than all intrinsic metrics.}
\label{fig:combined-corrs}
\end{figure*}

\subsection{Task Dimension}\label{sec:task_dim}
Table~\ref{tab:proxy_delta} presents the per-task comparison for generator selection. Visualisation can be found in Appendix~\ref{sec:appendix_add_visualisations}. The LLM parameter size heuristic proves effective for topic classification but performs poorly on sentiment analysis. This discrepancy arises because the heuristic identifies Llama 3 70B as the optimal LLM: while this model captures the topic task well, it struggles to generate high-quality data for sentiment analysis, the most diverse task (see Section~\ref{sec:data_tasks}).

By contrast, RoSE delivers consistently strong performance across tasks. It achieves the best results in both intent and sentiment tasks, ranking second in topic classification.

\subsection{Language Dimension}\label{sec:lang_dim}

Table~\ref{tab:metric_means} presents the per-language comparison of proxy metric performance for selecting an LLM generator. Visualisation can be found in the Appendix~\ref{sec:appendix_add_visualisations}. In two cases, RoSE is the second-best metric for selecting an optimal LLM generator (Indonesian and Welsh). However, in both cases, the gap in F1 performance is relatively small. 

For all the other cases, RoSE is the best proxy metric for selecting the optimal generator. RoSE is the best proxy metric for languages with various transcripts, working well for Telugu, Thai or Hebrew. We also note that no other proxy metrics work consistently well: the second-best proxy of LLM parameter size fails on languages such as German, Hebrew, Thai or Telugu.

\subsection{Proxy Metric Correlations with Downstream Model Performance}

Figure~\ref{fig:results_corrs} presents a forest plot of mean Pearson correlations between each proxy metric and downstream classifier performance on human-labelled test data, with error bars indicating confidence intervals. Our results highlight the limitations of existing intrinsic metrics. Measures such as silhouette score and token entropy often exhibit weak or even negative correlations with human-based performance, while type–token ratio and bigram diversity yield only small and unstable positive correlations. Heuristic proxies, such as dataset size and LLM parameter size, also fail to provide reliable guidance, as their correlations are low and the confidence intervals cross zero. In contrast, RoSE achieves a consistently strong correlation with downstream performance, with a mean Pearson correlation of around 0.6 and confidence intervals entirely above zero.

We also provide Rank correlations (via Kendal $\tau$) in Figure~\ref{fig:results_corrs_kendal} to investigate if we can rank the performance of LLMs as generators by using proxy metrics. RoSE is the only metric to achieve mild positive correlations, indicating that it consistently points in the right direction for LLM ranking as generators. However, the correlations in both cases are not near-perfect, indicating that the gap to optimal LLM selection and ranking remains. Our findings confirm prior observations that intrinsic metrics are not reliable indicators of synthetic data utility~\citep{cegin2024effectsdiversityincentivessample}.

\subsection{Excluding Largest LLM From Comparison}

The second-best metric in our main results is the LLM parameter size heuristic, which consistently selects Llama 3 70B. To test the viability of this metric and our own RoSE, we conducted an ablation study excluding Llama 3 70B and comparing the remaining five LLMs listed in Section~\ref{sec:models}, where the next largest model is Gemma 3 27B. The results, shown in Figure~\ref{fig:results_main_ablt}, demonstrate that RoSE remains a strong proxy, with only a -1.26\% F1 gap relative to the optimal LLM generator. In contrast, the next-best metric, average cosine distance, shows a larger gap of -3.62\%, while the parameter size heuristic drops to -5.78\%, performing worse than random selection. These findings highlight that RoSE continues to perform reliably even when competing LLMs are closer in parameter scale. 

\subsection{Performance on Varying Number of Candidate LLMs}
\begin{figure}[t!]
    \centering
    \includegraphics[width=7.7cm]{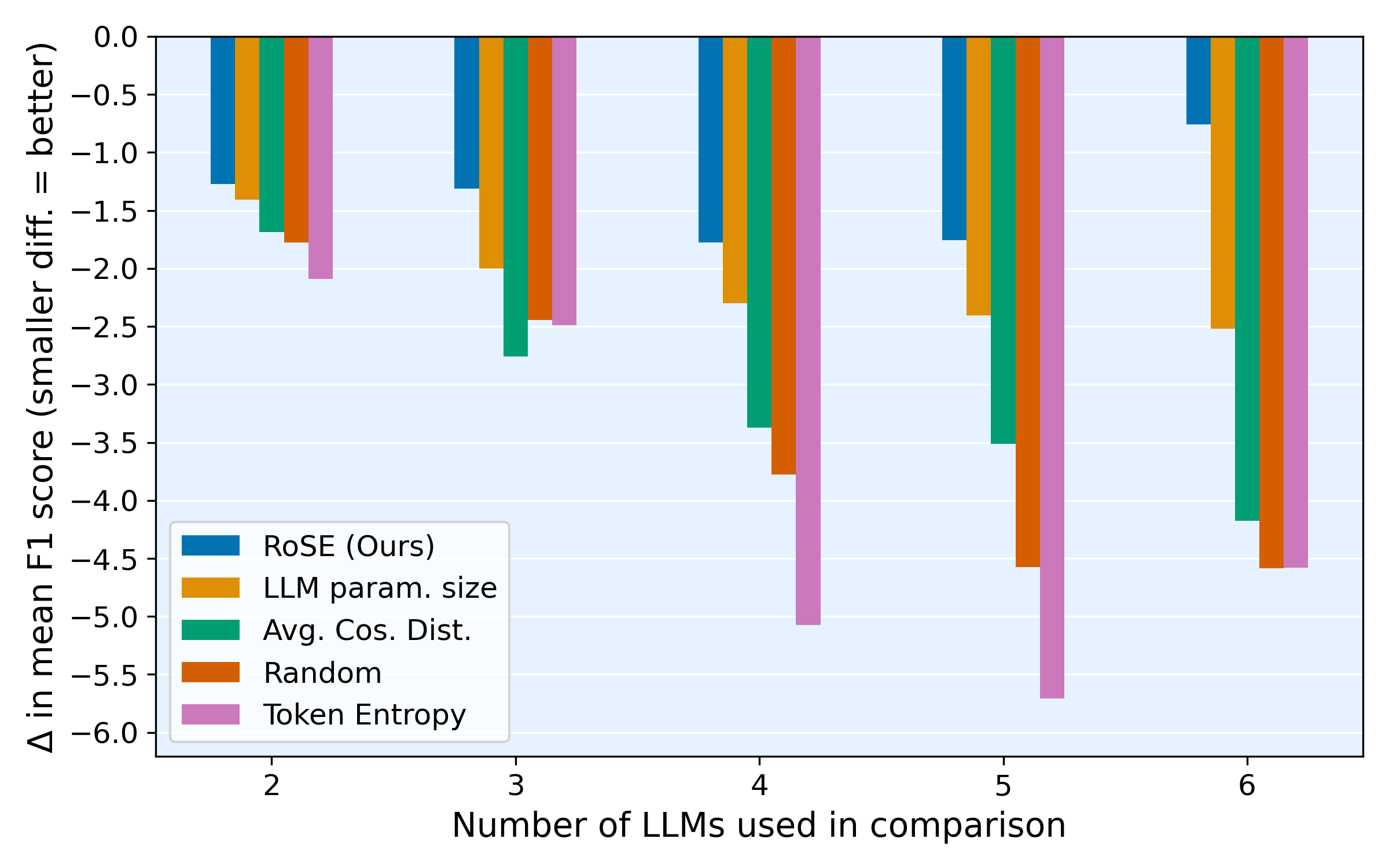}
    \caption{Comparison of a selection of proxy metrics for selecting the best LLM generator when comparing various combinations of LLMs (all combinations of up to 6 LLMs are considered). RoSE is the best proxy metric for a varying number of LLMs compared.}
    \label{fig:results_no_llms_compared}
\end{figure}

We evaluate how RoSE compares to other proxy metrics in selecting LLM generators across different numbers of models. Specifically, we consider all unique combinations of 2 to 6 LLMs. For the case of 6 LLMs, the results coincide with those shown in Figure~\ref{fig:results_main}. The outcomes for each number of LLMs are illustrated in Figure~\ref{fig:results_no_llms_compared}, highlighting RoSE alongside the strongest-performing alternative metrics. The visualisation breakdown per task can be found in Appendix~\ref{sec:appendix_varying_numb_of_cand_llms}.

Across all settings, RoSE consistently emerges as the best proxy metric for identifying the optimal LLM generator, with the smallest performance gap observed when comparing two models. Moreover, RoSE exhibits the most stable mean gap across varying numbers of LLMs, underscoring its predictive reliability across different model combinations.

\subsection{Cost Effectiveness of RoSE}\label{sec:coste_effective_rose}

Although RoSE demonstrates strong performance in selecting the optimal LLM generator, it is also computationally expensive. For each LLM under comparison, RoSE requires repeatedly training a smaller model and evaluating it on data generated by the other LLMs. In contrast, embedding-based approaches or simple heuristics, such as LLM parameter size, are far less costly to compute. An overview Table~\ref{tab:proxy_metrics_summary} can be found in Appendix~\ref{sec:appendix_add_visualisations} with relative computational costs and performance.

To explore how reducing the cost of RoSE affects its performance, we consider a simplified variant. Instead of evaluating the trained smaller model on test sets from all other LLMs, we randomly select a single alternative LLM to provide the evaluation set. This process is repeated for every LLM under comparison 10 times. We then assess whether this random selection strategy leads to a significant drop in RoSE’s ability to identify the optimal LLM generator. We consider selecting randomly 1 or a combination of various LLMs for up to all 6. The results are shown in Figure~\ref{fig:results_no_llms_used_comparison}. For per-task breakdown, see Appendix~\ref{sec:appendix_cost_effectivness_rose_per_task}.

\begin{figure}[t!]
    \centering
    \includegraphics[width=7.7cm]{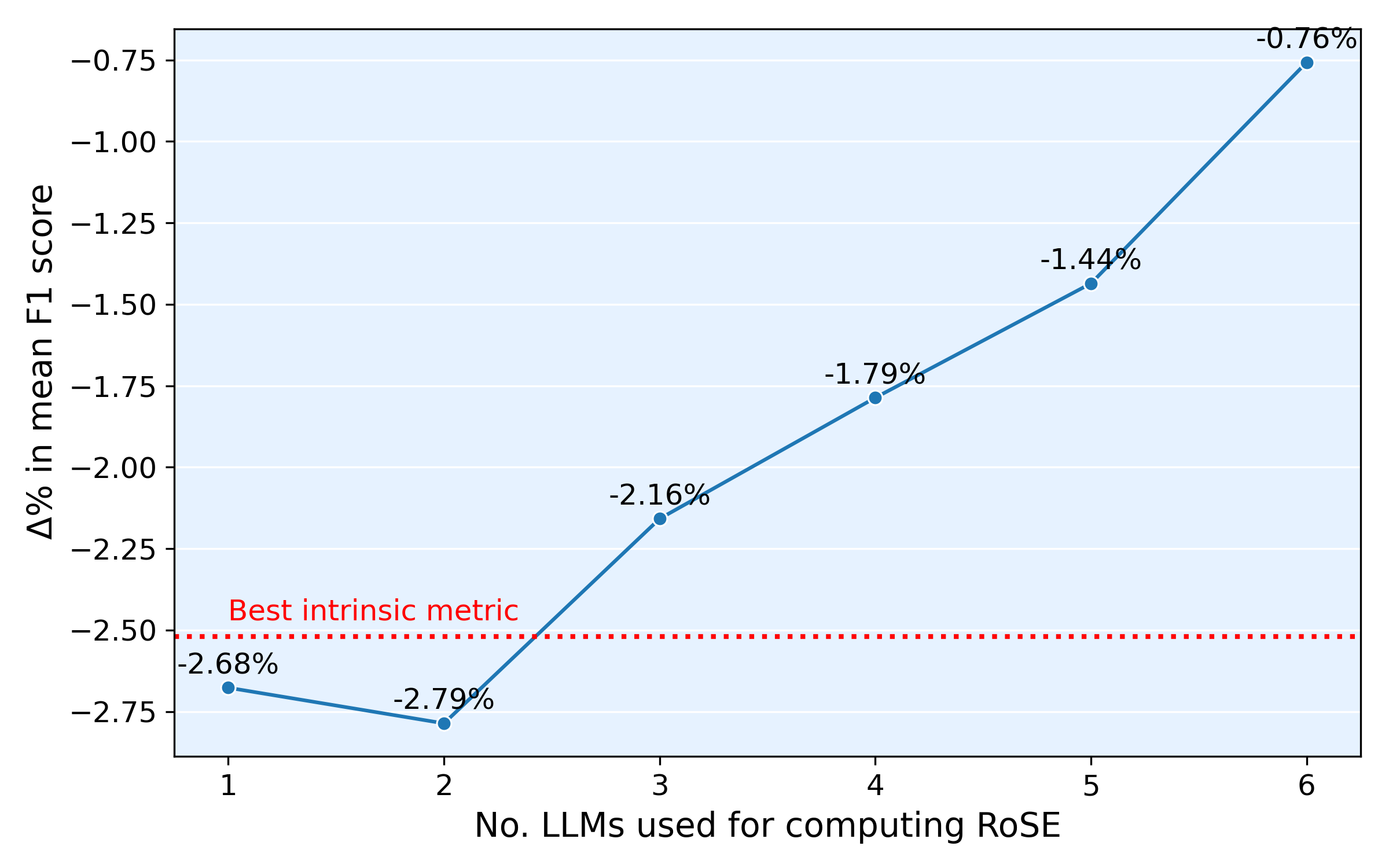}
    \caption{Number of randomly chosen LLMs used for computing RoSE and its effect on F1 score difference to optimal LLM generator selection. RoSE benefits from more LLMs being used during the evaluation of the downstream classifier during its computation.}
    \label{fig:results_no_llms_used_comparison}
\end{figure}

Our results show that RoSE outperforms all other proxy metrics with as few as three LLMs used for evaluation, surpassing the next best proxy of LLM parameter size. With the exception of the two-LLM case, the performance gap when using RoSE to select the optimal LLM generator decreases as more models are included. The combination of two randomly chosen LLMs for computing RoSE can lead to significant variability of the computed values by providing proxy performance that can be skewed. This suggests that RoSE benefits from a larger pool of LLMs during evaluation, providing a closer approximation to human performance.

We further examine whether data from certain LLMs consistently degraded RoSE’s effectiveness. While some models performed poorly in specific languages (e.g., Qwen 2.5 on Hebrew), we did not find any single LLM whose outputs were consistently detrimental. Thus, excluding specific LLMs from RoSE computation is not necessary.

\section{The Role of Prompt Examples in RoSE}\label{sec:no_icl_used}

As described in Section~\ref{sec:experiments}, we follow the strategy of~\cite{anikina2025rigorousevaluationllmdata}, prompting LLMs with 10 human examples during data generation. This in-context approach has been shown to yield the best downstream performance.  

To examine how these examples affect RoSE’s predictive strength, we also evaluate a zero-shot variant, where prompts contain no examples and only LLM self-revision is applied, similar to the revision-only strategy of~\cite{anikina2025rigorousevaluationllmdata}. We denote our metric in this setting as \textbf{RoSE-Z}.  

\begin{figure}[t!]
    \centering
    \includegraphics[width=7.7cm]{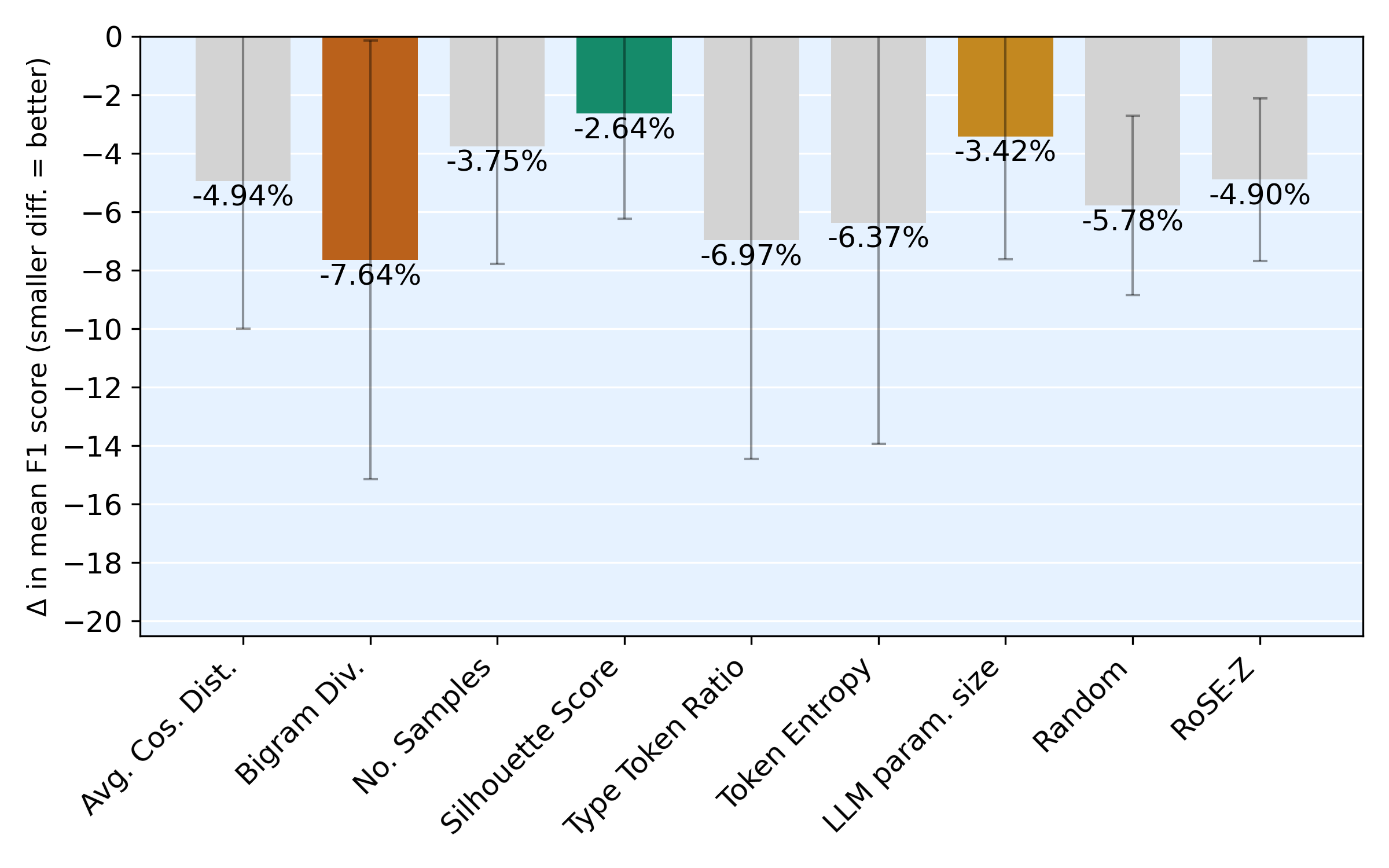}
    \caption{Comparison of proxy metrics for selecting the best LLM generator when generation setup without examples is used (zero-shot). RoSE-Z in this setup struggles to identify the optimal LLM generator, underlining the importance of including in-context examples in the data generation process.}
    \label{fig:results_rose_z}
\end{figure}

\textbf{We note that classifiers trained on zero-shot data perform substantially worse on human test sets, with an average drop of 8.45\% in F1}, consistent with prior findings~\cite{piedboeuf-langlais-2023-chatgpt, anikina2025rigorousevaluationllmdata}. Moreover, RoSE-Z is less effective at identifying the optimal LLM generator: while better than random selection, three alternative metrics outperform it (see Figure~\ref{fig:results_rose_z}).  

These results suggest that RoSE’s strong predictive power depends on the inclusion of human examples in prompts. RoSE’s effectiveness as a proxy metric seems to arise from its ability to approximate human-like data generation, making it a reliable proxy for extrinsic evaluation. This is in line with previous research that identified including examples during generation as beneficial for the downstream model performance and that such data resembles human data more closely~\cite{anikina2025rigorousevaluationllmdata, cegin2024effectsdiversityincentivessample}.

\section{Conclusion}

In this work, we proposed RoSE (Round-robin Synthetic data Evaluation) as a principled proxy metric for selecting the most effective LLM generator in scenarios where human-labelled test sets are unavailable. Our extensive experiments across 11 languages and three diverse classification tasks demonstrate that RoSE consistently outperforms intrinsic metrics and heuristic baselines, identifying the optimal LLM most consistently and achieving an average F1 gap of just 0.76\% for chosen LLM generators by this proxy. Beyond its predictive strength, RoSE proves reliable across tasks, languages, and even when the number of candidate LLMs is small, making it a practical tool for low-resource settings. Overall, RoSE represents a step toward the reliable and cost-effective selection of LLM generators in data-scarce scenarios.

\section*{Acknowledgements}

This work was partially funded by the European Union under the project lorAI - Low Resource Artificial Intelligence, GA No. 101136646, and by NextGenerationEU through the Recovery and Resilience Plan for Slovakia under the projects No. 09I01-03-V03-00020 (AI-Auditology), and No. 09I01-03-V04-00068 (GEPERO).

This work was supported by the Ministry of Education, Youth and Sports of the Czech Republic through the e-INFRA CZ (ID:90254).

\section*{Limitations}

Due to the scope of this study, we limited ourselves to 6 different LLMs used. We tried to mitigate this limitation by including LLMs of various sizes and from various LLM families. We additionally used only 11 languages and 3 tasks due to resource constraints and the availability of evaluation data for the low-resource languages.

We limited the number of used labels for the MASSIVE dataset to the 10 most common intents to avoid many conflating factors that can be potentially caused by the semantic overlaps in label descriptions.

We acknowledge that the extent to which data contamination affects RoSE's performance is unknown. The data on which LLMs were trained is not fully disclosed, which might be reflected in the disparities between the LLMs’ performance when generating data in different languages and domains.

A limitation of the RoSE method is its requirement of needing a few (10 per label) human examples for the generation process. As we theorise in the paper, the generated data is then more similar to the human data we are trying to replace during evaluation (and computation of RoSE). Without it, RoSE performs significantly worse, but so do downstream models trained on data generated without human examples included. As such, a small number of human examples is still essential, both for RoSE performance and the general good performance of LLMs as generators.

\section*{Ethical Considerations}
Based on a thorough ethical assessment performed on the basis of intra-institutional ethical guidelines and checklists tailored to the use of data and algorithms, we see no ethical concerns pertaining directly to the conduct of this research. Although the production of new data through LLMs bears several risks, such as the introduction of biases, the small size of the produced dataset, sufficient for experimentation, is, at the same time, insufficient for any major machine learning endeavours where such biases could be transferred.

We follow the license terms for all the models and datasets we used (such as the one required for the use of the Llama-3 model) – all models and datasets allow their use as part of the research.

\bibliography{custom}
\clearpage
\appendix

\section{Appendix}

\subsection{Language Abbreviations}\label{sec:lang_abbreviations}
\begin{table}[h]
\centering
\footnotesize
\begin{tabular}{ll}
\toprule
Code &  Language \\
\midrule
    az & Azerbaijani \\
    cy & Welsh \\
    de & German \\
    en & English \\
    he & Hebrew \\
    id & Indonesian \\
    ro & Romanian \\
    sl & Slovenian \\
    sw & Swahili \\
    te & Telugu \\
    th & Thai \\
\bottomrule
\end{tabular}
\caption{Language abbreviations.}
\label{tab:lang_abbreviations}
\end{table}

\begin{figure}[h]
    \centering
    \includegraphics[width=7.7cm]{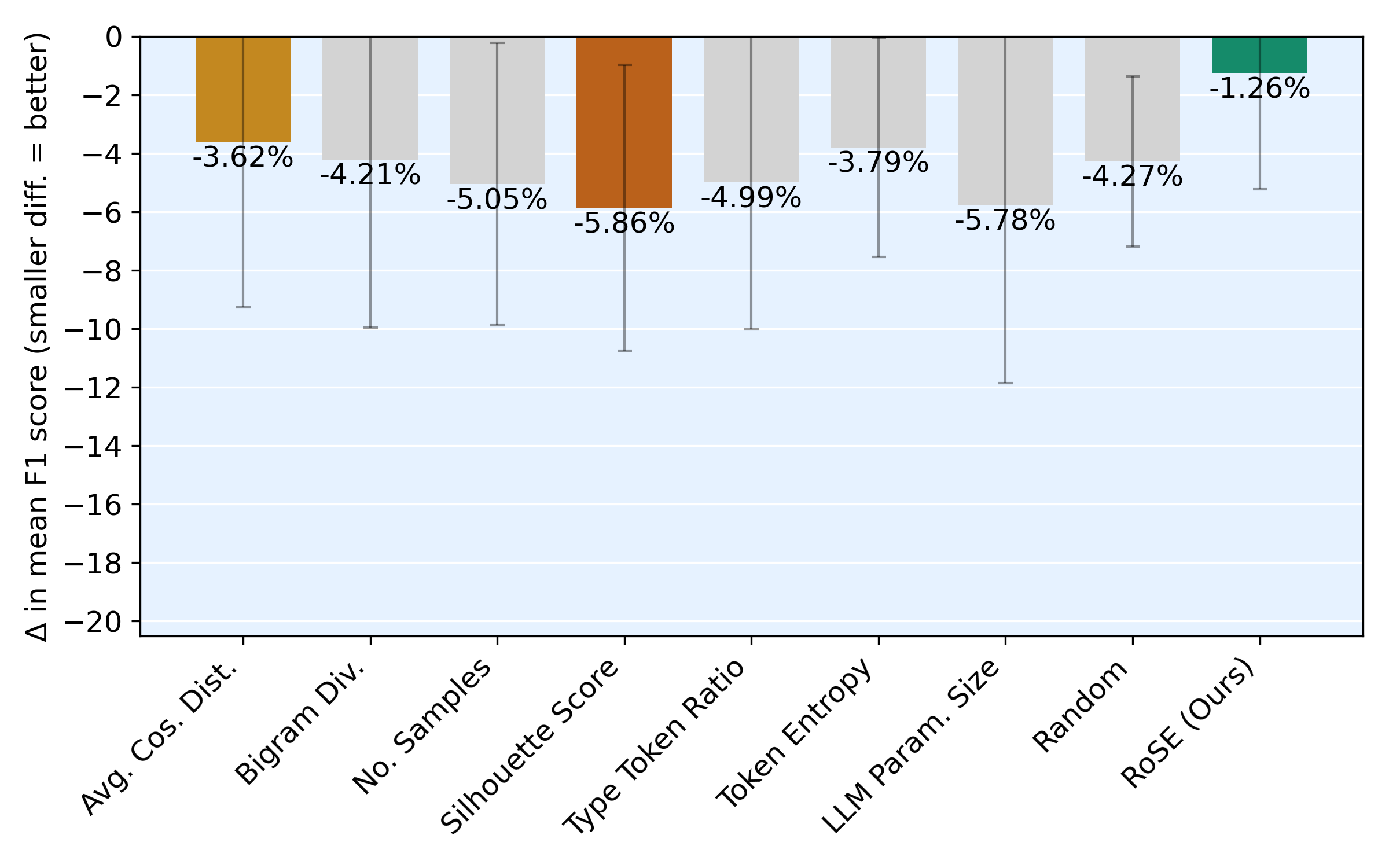}
    \caption{Comparison of proxy metrics for selecting the best LLM generator without considering Llama 3 70B in the comparison. Bars show the average gap in mean F1 score to the optimal human-based evaluation (smaller is better). The best metric is green, the second best is orange, and the worst is red.}
    \label{fig:results_main_ablt}
\end{figure}

\subsection{Computational Resources}\label{sec:computational_resources}
All generation experiments were done using vllm \cite{kwon2023efficient} for efficient inference. For a single language-model combination, it takes approximately 2 hours to generate the data for intent recognition, 1 hour 20 minutes for topic classification and 23 minutes for sentiment classification. We use a single H100 GPU to generate the samples. Thus, for 11 languages, 6 LLMs and 3 different tasks, it takes around 180 GPU hours to generate all the data.

The fine-tuning experiments were also performed on an H100 GPU. All the finetuning experiments took approximately an additional 200 GPU hours.

\subsection{Downstream Fine-tuning of XLM-R}\label{sec:appendix_downstream_finetuning}
For the downstream evaluation, we fine-tune the XLM-R \cite{DBLP:journals/corr/abs-1911-02116} \textit{FacebookAI/xlm-roberta-base} model for 50 epochs with a batch size of 16 and employ early stopping with a patience of 5 epochs to prevent overfitting. We perform hyperparameter optimisation to determine the optimal learning rate and set it to 1e-5. \textit{AdamW} is used as an optimiser. We balance the generated datasets to have the same number of samples per label. We normalise all inputs by converting them to lowercase and removing punctuation.

\subsection{Prompt Templates and Generation Details}\label{sec:appendix_prompt_templates}

For generating synthetic data, we used this general prompt template with examples included:

\textit{"Please create 6 different \{task\_text\_type\} in the \{language\} language, separated by the {sep\_token} token. The \{task\_text\_type\} should be about the \{label\}. Note that some examples from the dataset look as follows: Examples: \{examples\}. Output only the text in \{language\} and nothing else! Do not number the texts!"}

The \textit{task\_text\_type} placeholder had values based on tasks, either semantic analysis, intent recognition or topic classification. The \textit{sep\_token} was represented as "---". The \textit{label} placeholder was replaced for each task and label with an LLM-generated explanation of that label based on randomly preselected human examples. The \textit{examples} placeholder contained the human examples for that given label being generated. For the RoSE-Z generation, this was excluded, and no examples were given to the LLM generator.

Sampling parameters used for vllm generation were: \textit{temperature=0.7,  top\_p=0.9, max\_tokens=4096, repetition\_penalty=1.2}. We collected 10 generated samples per inference run for increased efficiency and performed cleaning of the generated data. We excluded duplicates and collected until 100 unique texts were generated.

Specific versions of LLMs used for generations were: Llama-3-70b-instruct~\footnote{\url{https://huggingface.co/TechxGenus/Meta-Llama-3-70B-Instruct-GPTQ}}, Llama-3-8b-instruct~\footnote{\url{https://huggingface.co/TechxGenus/Meta-Llama-3-8B-Instruct-GPTQ}},  Qwen2.5-14B-Instruct~\footnote{\url{https://huggingface.co/Qwen/Qwen2.5-14B-Instruct}}, Magistral-Small-2509~\footnote{\url{https://huggingface.co/mistralai/Magistral-Small-2509}}, gemma-3-27b-it~\footnote{\url{https://huggingface.co/google/gemma-3-27b-it}}, gemma-3-4b-it~\footnote{\url{https://huggingface.co/google/gemma-3-4b-it}}.

\subsection{Performance on Varying Number of Candidate LLMs: Breakdown Per Task}\label{sec:appendix_varying_numb_of_cand_llms}

\begin{figure*}[t!]
\centering
\begin{subfigure}{0.32\textwidth}
\centering
\includegraphics[width = \textwidth]{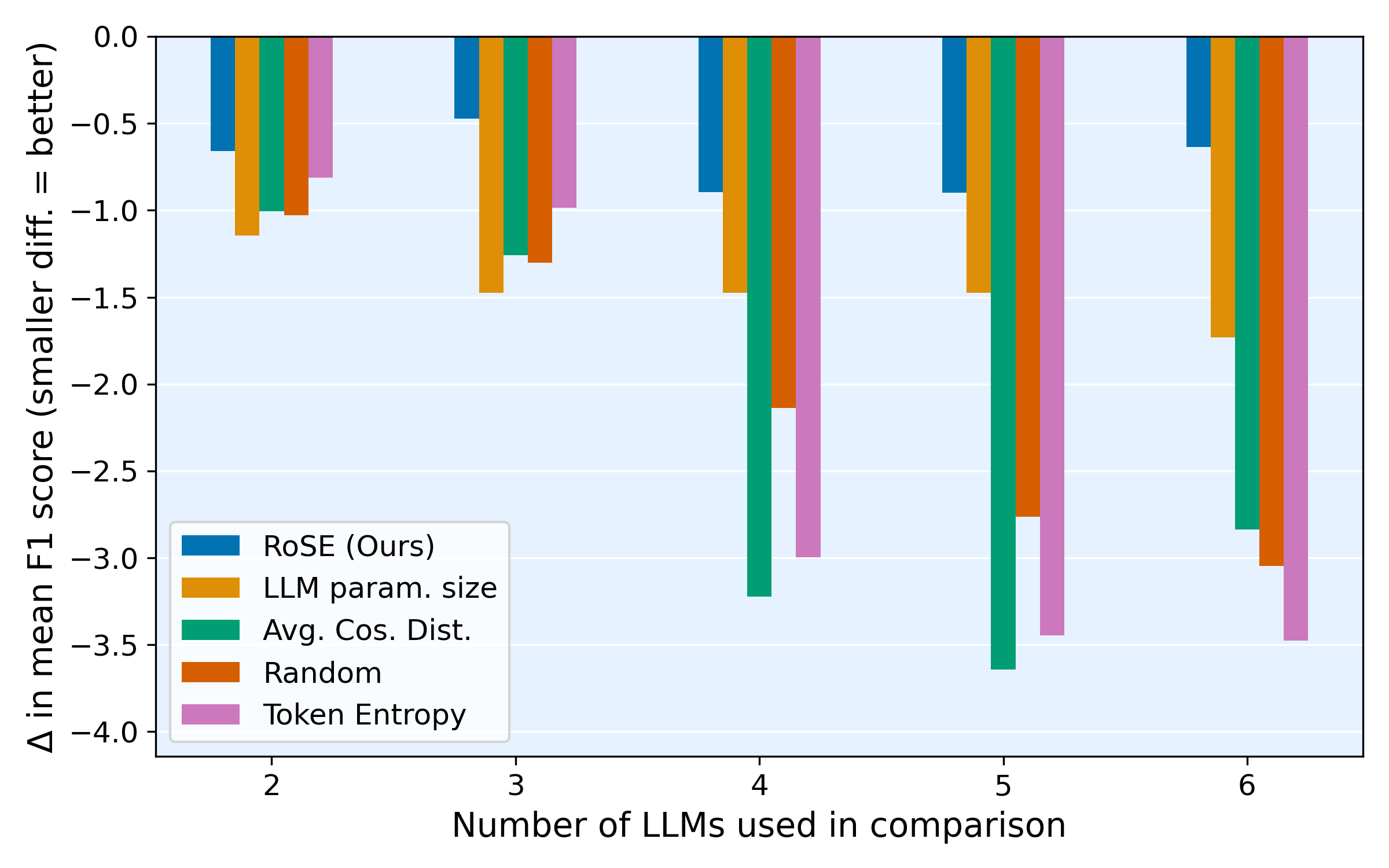}
\caption{Intent recognition.}
\label{fig:intent_task_barplots_number_llms}
\end{subfigure}
\begin{subfigure}{0.32\textwidth}
\centering
\includegraphics[width = \textwidth]{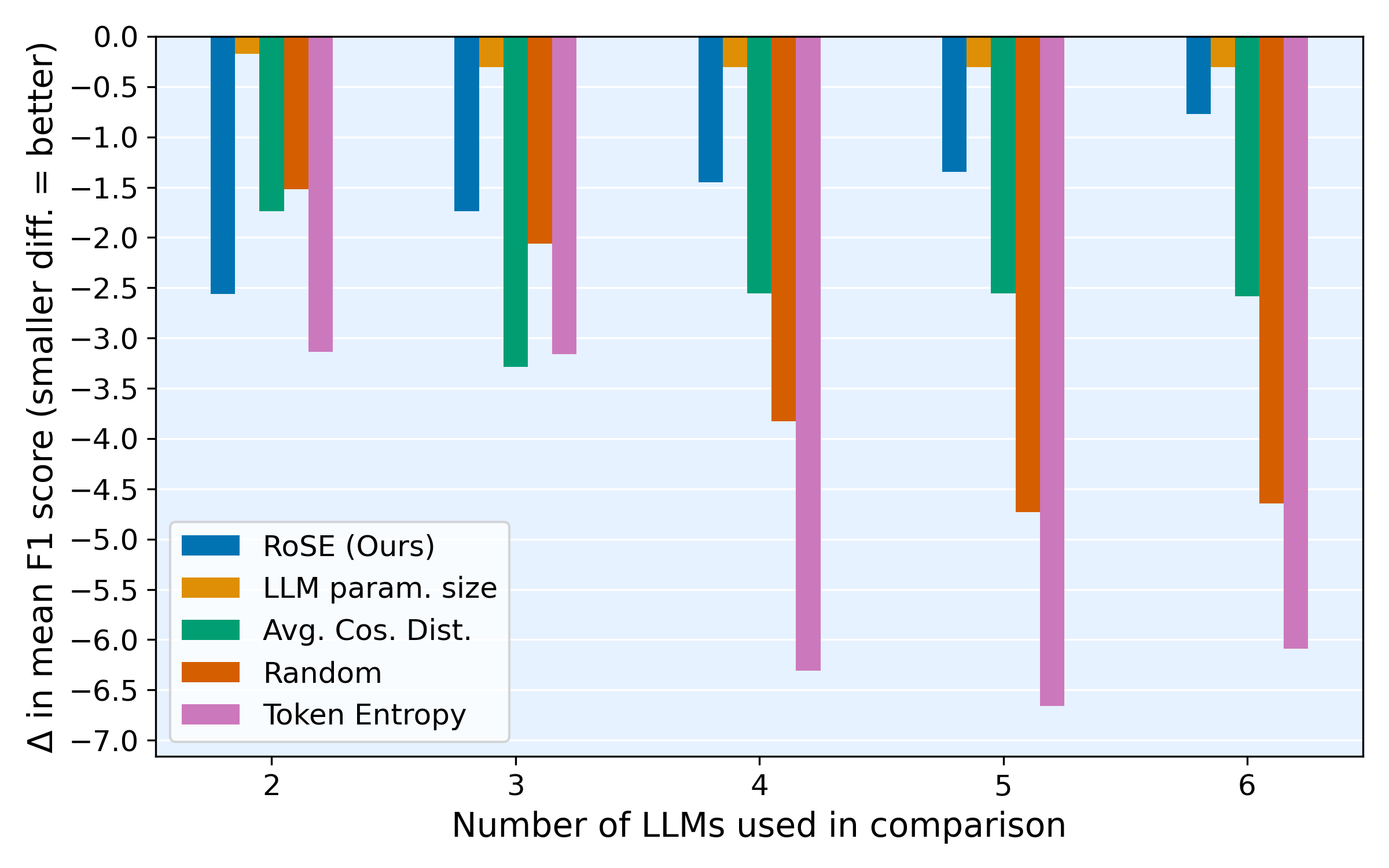}
\caption{Topic classification.}
\label{fig:topic_task_barplots_number_llms}
\end{subfigure}
\begin{subfigure}{0.32\textwidth}
\centering
\includegraphics[width = \textwidth]{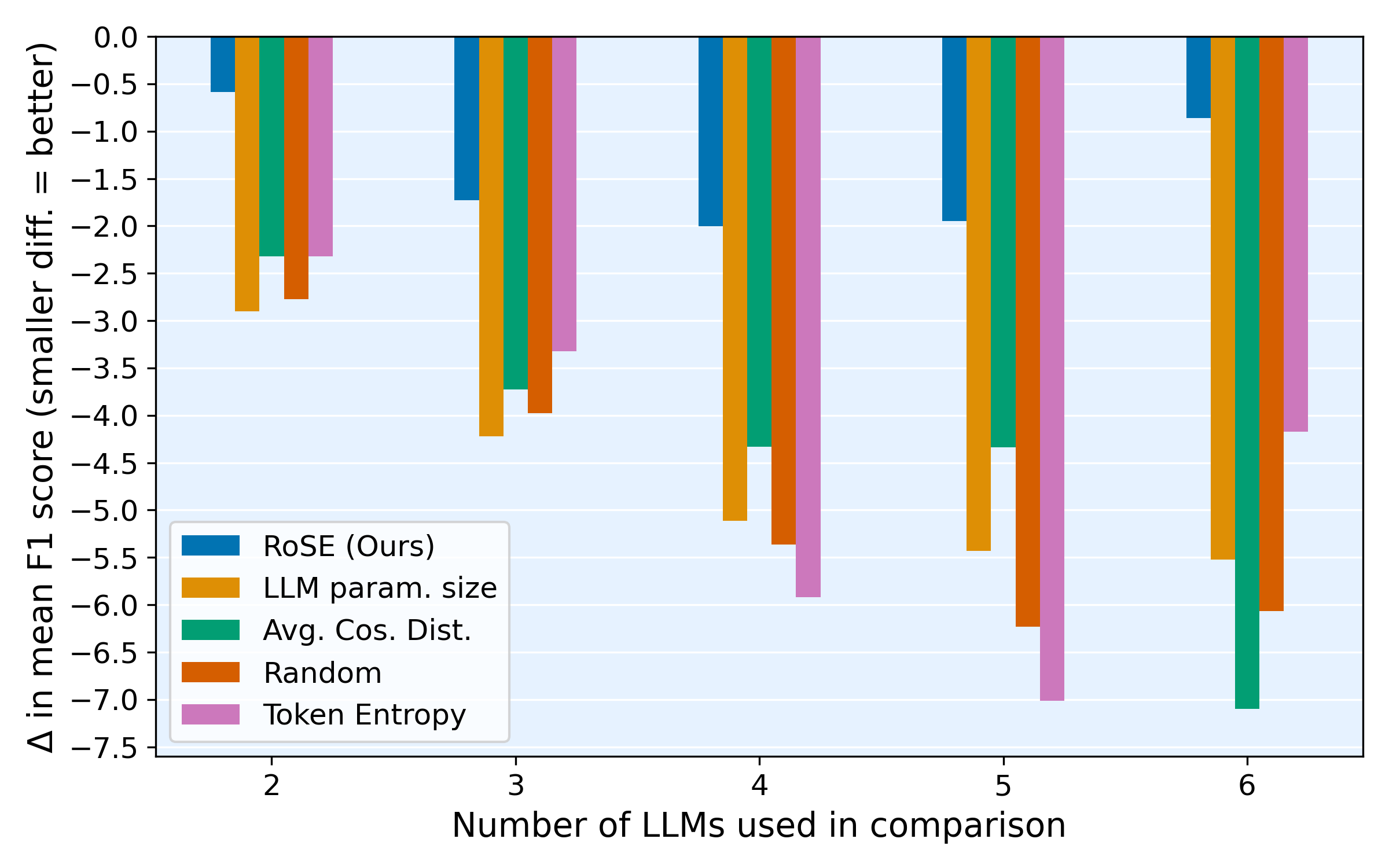}
\caption{Sentiment analysis.}
\label{fig:sentiment_task_barplots_number_llms}
\end{subfigure}
\caption{Comparison per-task of a selection of proxy metrics for selecting the best LLM generator when comparing various combinations of LLMs.}
\label{fig:combined-per-task-barplot-number-llms}
\end{figure*}

We provide a breakdown per task of the selected best-performing proxy metrics when using a varying number of LLMs in comparisons. Our results in Figure~\ref{fig:combined-per-task-barplot-number-llms} show that RoSE remains the best performing metric except for Topic classification, where LLM parameter size's performance can be explained due to Llama-3 70B's impressive data generation capabilities for this task.

\subsection{Cost Effectiveness of RoSE: Breakdown Per Task}\label{sec:appendix_cost_effectivness_rose_per_task}

\begin{figure*}[t!]
\centering
\begin{subfigure}{0.32\textwidth}
\centering
\includegraphics[width = \textwidth]{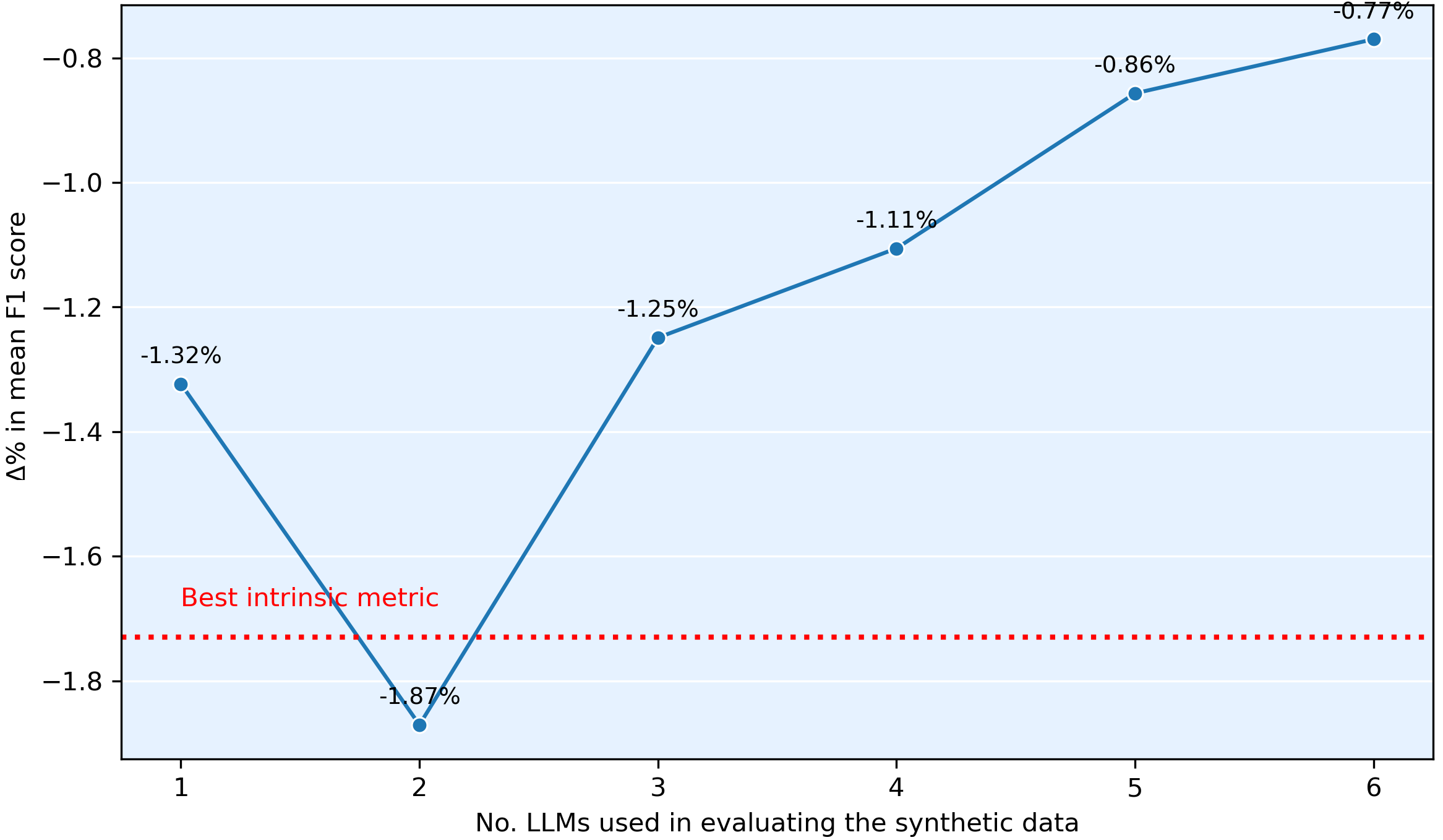}
\caption{Intent recognition.}
\label{fig:intent_task_diffs}
\end{subfigure}
\begin{subfigure}{0.32\textwidth}
\centering
\includegraphics[width = \textwidth]{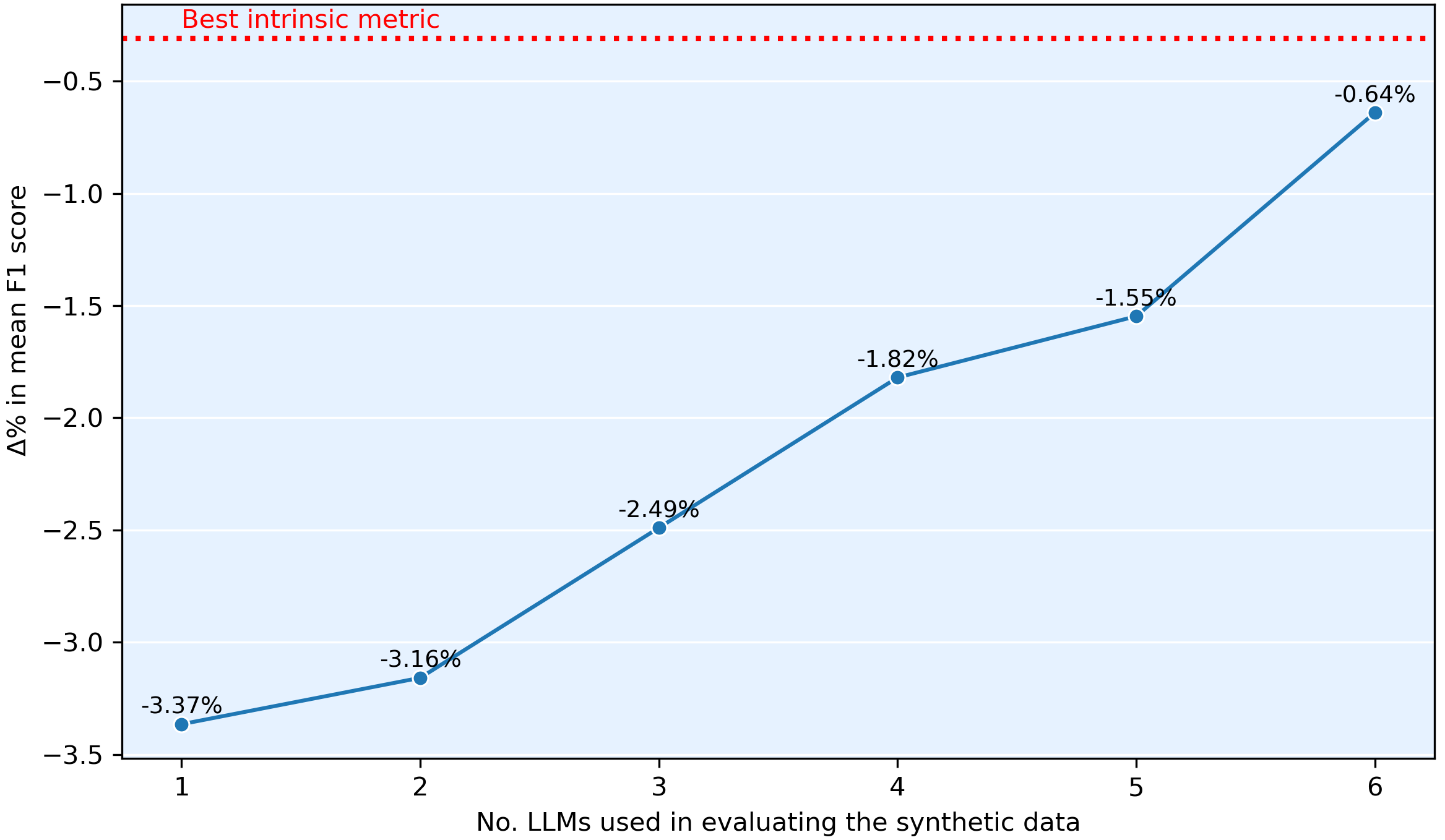}
\caption{Topic classification.}
\label{fig:topic_task_diffs}
\end{subfigure}
\begin{subfigure}{0.32\textwidth}
\centering
\includegraphics[width = \textwidth]{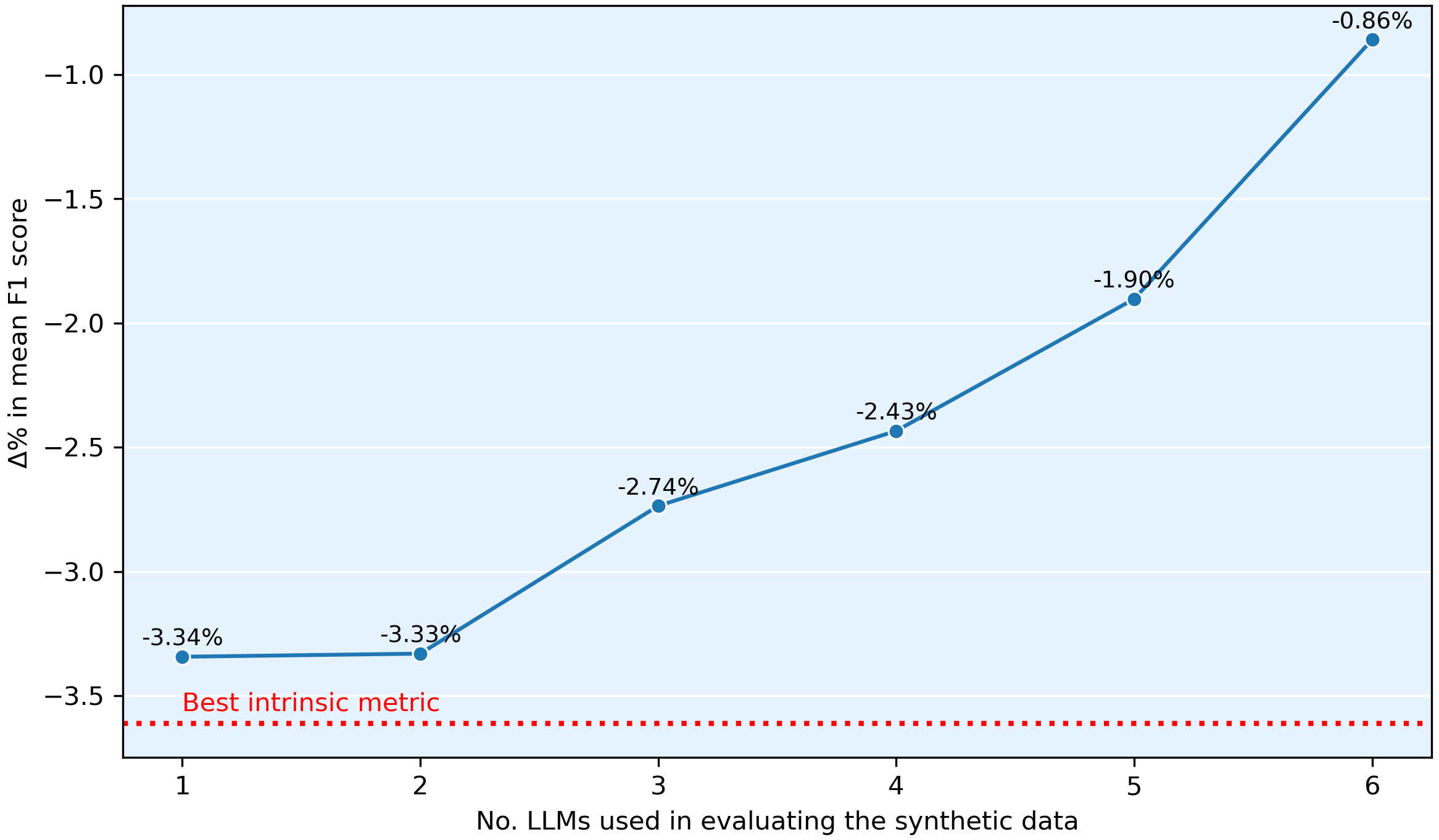}
\caption{Sentiment analysis.}
\label{fig:sentiment_task_diffs}
\end{subfigure}
\caption{Number of randomly chosen LLMs used for computing RoSE and its effect on F1 score difference to optimal LLM generator selection. RoSE benefits from more LLMs being used during the evaluation of the downstream classifier during its computation. For all tasks, the performance increases from 3 or more LLMs used for computing RoSE.}
\label{fig:combined-per-task-diffs-per-llm}
\end{figure*}

We provide a cost-effective breakdown of RoSE per task in Figure~\ref{fig:combined-per-task-diffs-per-llm}. Similar to findings in Section~\ref{sec:coste_effective_rose}, the performance of RoSE in identifying a good LLM generator increases with more LLMs' synthetic data being used during the computation of RoSE. The only exception is when using 2 randomly chosen LLMs for the intent recognition task, where a sudden drop is present. The combination of 2 randomly chosen LLMs for computing RoSE can lead to significant variability of the computed values by providing proxy performance that can be skewed. Including 3 or more LLMs for computing RoSE is thus essential for the performance of these proxy metrics.

\begin{table*}[t!]
\centering
\small
\begin{tabular}{lccccccccccc}
\toprule
LLM & az & cy & he & th & sw & sl & en & de & id & ro & te \\
\midrule
Magistral-Small & 66.90 & 47.19 & 70.62 & 69.78 & 48.48 & 79.53 & 77.24 & 66.20 & 83.26 & 82.06 & 68.88 \\
Meta-Llama-3-70B & 65.98 & 57.95 & 69.81 & 76.24 & 55.63 & 75.77 & 72.97 & 59.58 & 87.82 & 85.79 & 53.32 \\
Meta-Llama-3-8B & 70.01 & 45.28 & 69.98 & 65.15 & 60.48 & 76.92 & 79.26 & 69.36 & 81.63 & 80.23 & 70.26 \\
Qwen2.5-14B & 65.43 & 51.46 & 64.03 & 77.27 & 55.48 & 61.04 & 68.01 & 68.77 & 86.36 & 81.49 & 73.02 \\
gemma-3-27b & 64.44 & 36.72 & 77.17 & 69.48 & 47.69 & 77.65 & 63.41 & 69.36 & 88.71 & 77.04 & 75.61 \\
gemma-3-4b & 69.53 & 52.95 & 73.78 & 62.40 & 46.80 & 77.30 & 64.31 & 63.31 & 89.18 & 76.02 & 75.52 \\
\bottomrule
\end{tabular}
\caption{Mean F1 scores of XLM-R finetuned on data generated from each LLM via RoSE ICL generation setup per language for the sentiment analysis task on human test data. Higher is better.}
\label{tab:human_f1_table_sentiment}
\end{table*}

\begin{table*}[t!]
\centering
\small
\begin{tabular}{lccccccccccc}
\toprule
LLM & az & cy & he & th & sw & sl & en & de & id & ro & te \\
\midrule
Magistral-Small & 67.89 & 65.40 & 68.83 & 78.70 & 62.26 & 76.24 & 80.24 & 78.80 & 75.85 & 74.33 & 67.45 \\
Meta-Llama-3-70B & 76.56 & 65.47 & 72.86 & 76.86 & 65.86 & 80.33 & 80.15 & 81.07 & 81.06 & 78.38 & 69.97 \\
Meta-Llama-3-8B & 72.14 & 64.68 & 71.40 & 74.62 & 65.24 & 79.27 & 80.39 & 78.88 & 78.19 & 79.67 & 54.39 \\
Qwen2.5-14B & 73.22 & 65.09 & 46.57 & 73.33 & 61.66 & 76.55 & 78.39 & 76.70 & 73.74 & 76.02 & 62.63 \\
gemma-3-27b & 70.05 & 60.83 & 63.42 & 69.95 & 57.52 & 72.17 & 73.23 & 74.11 & 72.22 & 70.14 & 62.62 \\
gemma-3-4b & 66.06 & 52.17 & 65.65 & 71.21 & 58.89 & 69.92 & 70.59 & 74.53 & 71.44 & 71.12 & 62.42 \\
\bottomrule
\end{tabular}
\caption{Mean F1 scores of XLM-R finetuned on data generated from each LLM via RoSE ICL generation setup per language for the topic classification task on human test data. Higher is better.}
\label{tab:human_f1_table_topic}
\end{table*}

\begin{table*}[t!]
\centering
\small
\begin{tabular}{lccccccccccc}
\toprule
LLM & az & cy & he & th & sw & sl & en & de & id & ro & te \\
\midrule
Magistral-Small & 86.97 & 78.09 & 84.74 & 90.98 & 75.78 & 91.29 & 90.90 & 84.42 & 88.59 & 89.63 & 83.11 \\
Meta-Llama-3-70B & 84.93 & 75.77 & 87.21 & 85.10 & 73.64 & 91.12 & 92.58 & 85.99 & 90.82 & 89.58 & 84.07 \\
Meta-Llama-3-8B & 84.00 & 76.43 & 84.00 & 90.39 & 73.27 & 91.37 & 92.71 & 87.31 & 92.89 & 88.78 & 82.85 \\
Qwen2.5-14B & 85.01 & 65.29 & 82.16 & 87.90 & 65.31 & 90.09 & 91.05 & 89.14 & 91.93 & 90.61 & 82.09 \\
gemma-3-27b & 82.82 & 65.70 & 83.63 & 86.86 & 73.26 & 84.67 & 80.31 & 87.13 & 90.28 & 87.20 & 81.41 \\
gemma-3-4b & 82.14 & 67.23 & 85.13 & 87.29 & 71.05 & 88.03 & 91.15 & 80.27 & 92.43 & 86.92 & 80.44 \\
\bottomrule
\end{tabular}
\caption{Mean F1 scores of XLM-R finetuned on data generated from each LLM via RoSE ICL generation setup per language for the intent recognition task on human test data. Higher is better.}
\label{tab:human_f1_table_intent}
\end{table*}

\begin{table*}[t!]
\centering
%\small
\begin{tabular}{lc p{8cm}}
\toprule
Proxy Metric & Computational Cost & Performance \\
\midrule
Type--Token Ratio (TTR) & Low & Weak, low/unstable correlations with human performance \\
Average Cosine Distance & Medium & Moderate, sometimes effective but inconsistent \\
Bigram Diversity & Low & Weak, small or negative correlations \\
Number of Valid Samples & Low--Medium & Weak to moderate, unstable across tasks/languages \\
Silhouette Score & Medium & Weak, often a negative correlation with human data \\
Token Entropy & Low & Weak, small and inconsistent effects \\
LLM Parameter Size & Very Low & Second-best, strong for some tasks (topic), poor for others (sentiment) \\
Random Selection & Negligible & Baseline, consistently poor \\
RoSE & Highest & Best overall; smallest F1 gap to optimal, only metric with positive correlations \\
\bottomrule
\end{tabular}
\caption{Comparison of proxy metrics for selecting the best LLM generator. Costs are relative; performance is summarised based on overall results across languages and tasks.}
\label{tab:proxy_metrics_summary}
\end{table*}

\subsection{Additional Visualisations and Tables}\label{sec:appendix_add_visualisations}

\begin{figure*}[h!]
\centering
\begin{subfigure}{0.32\textwidth}
\centering
\includegraphics[width = \textwidth]{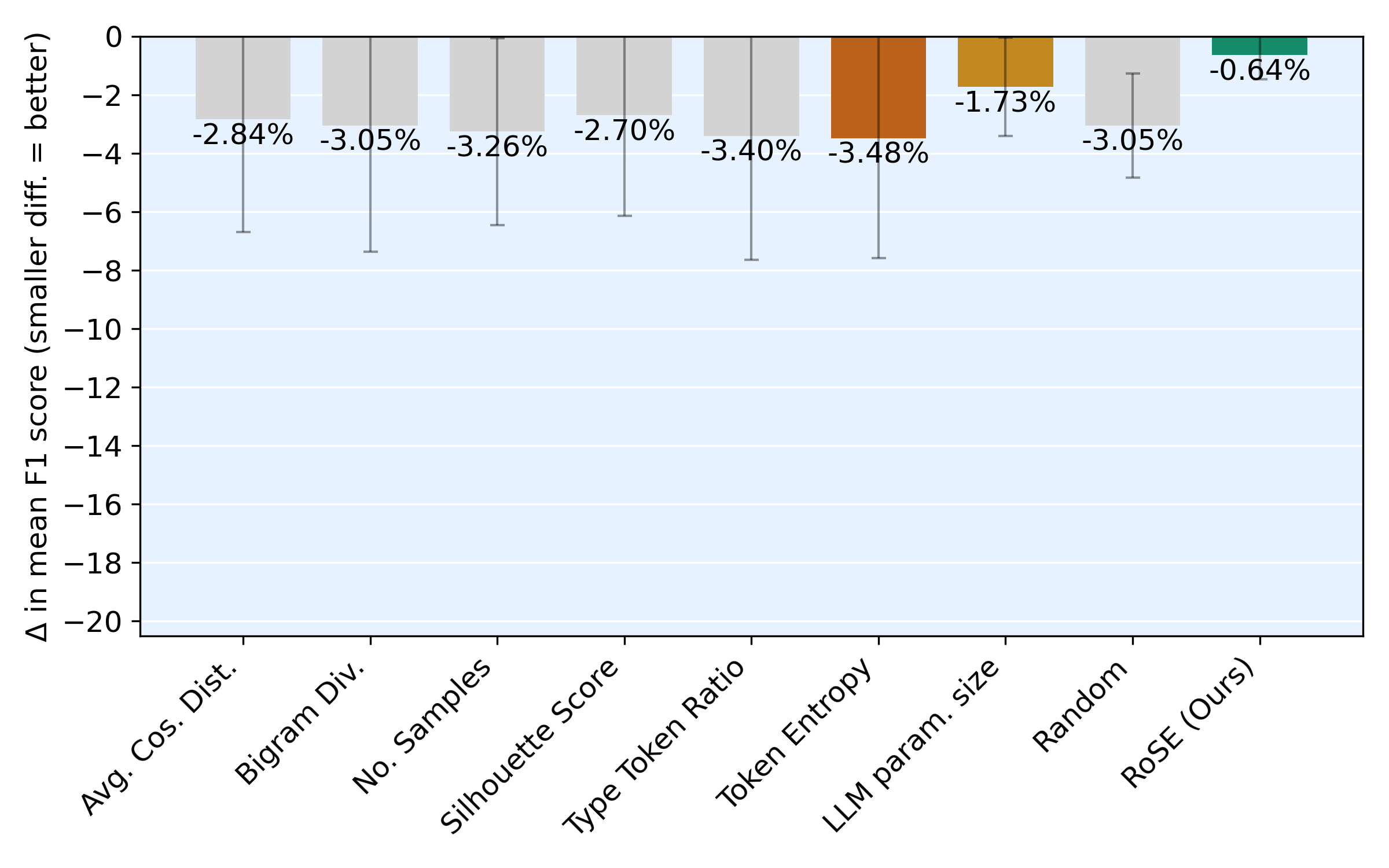}
\caption{Intent recognition.}
\label{fig:intent_task}
\end{subfigure}
\begin{subfigure}{0.32\textwidth}
\centering
\includegraphics[width = \textwidth]{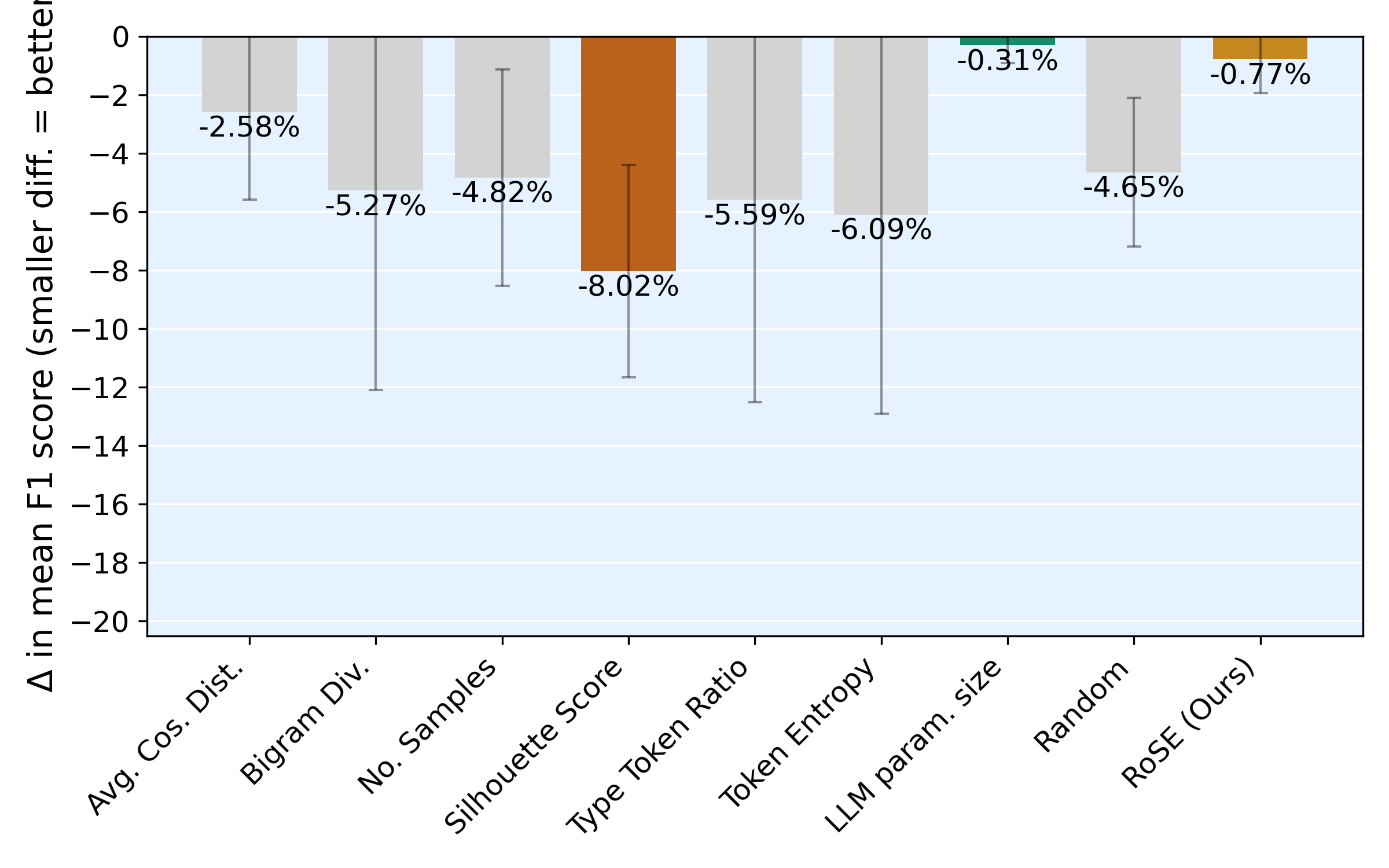}
\caption{Topic classification.}
\label{fig:topic_task}
\end{subfigure}
\begin{subfigure}{0.32\textwidth}
\centering
\includegraphics[width = \textwidth]{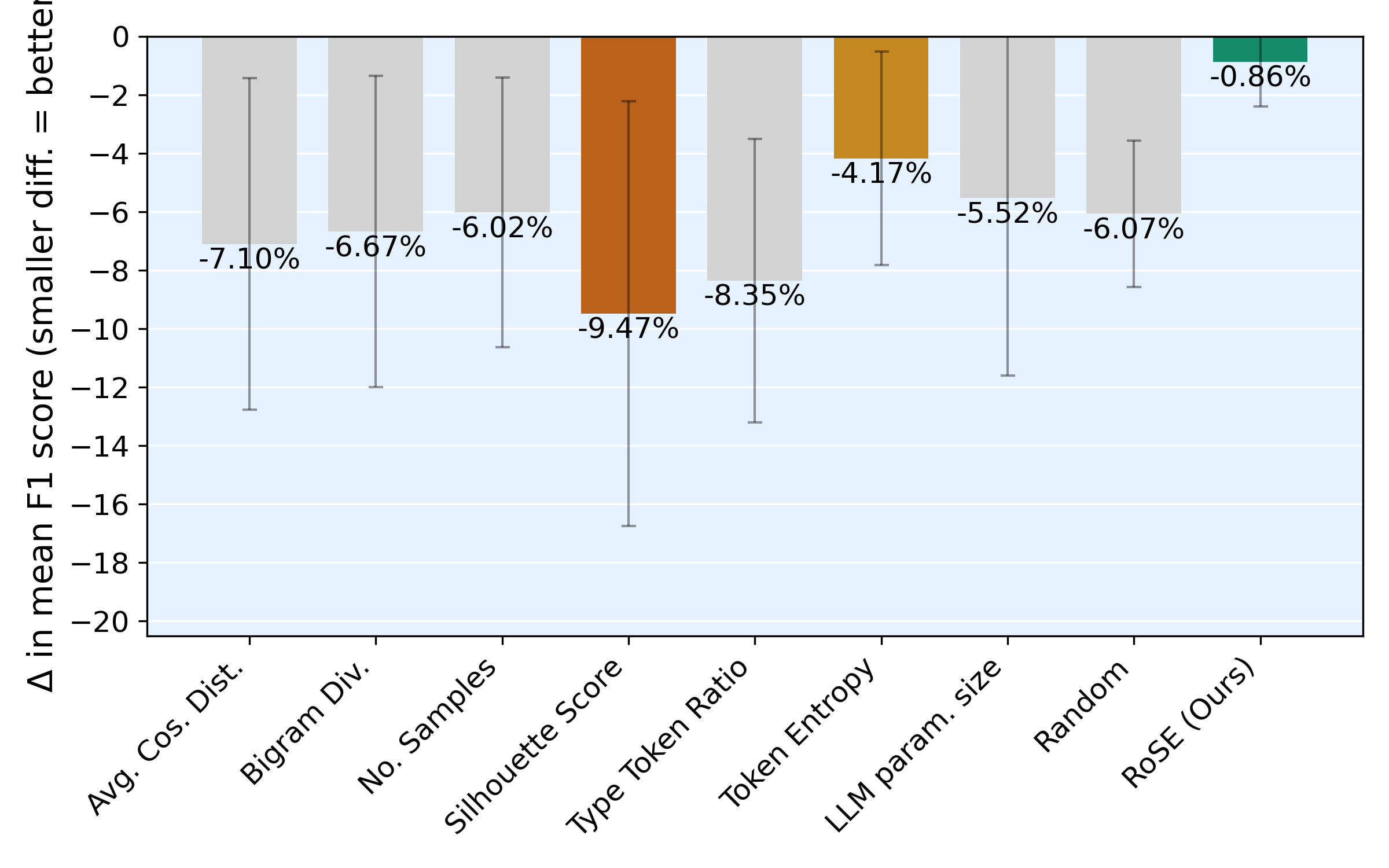}
\caption{Sentiment analysis.}
\label{fig:sentiment_task}
\end{subfigure}
\caption{Comparison of proxy metrics for selecting the best LLM generator per task.  Bars show the average gap in mean F1 score for models trained on the best generator selected by metrics vs. the optimal generator (smaller is better). The best metric is green, the second best is orange, and the worst is red. RoSE performs well across all tasks as either the best or the second-best proxy metric.}
\label{fig:combined-per-task}
\end{figure*}

\begin{figure*}[h!]
\centering
\begin{subfigure}{0.32\textwidth}
\centering
\includegraphics[width = \textwidth]{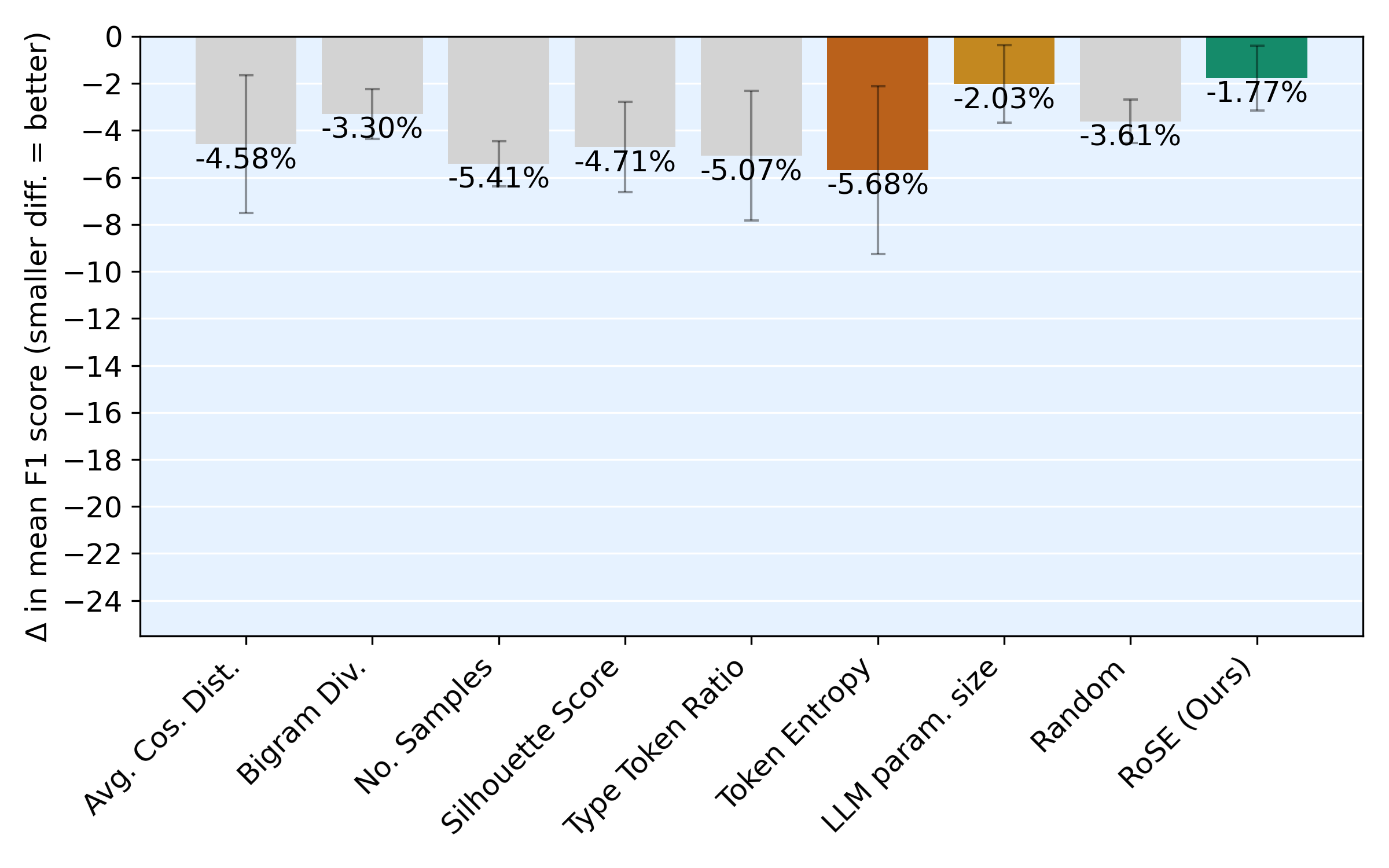}
\caption{Azerbaijani.}
\label{fig:per_lang_az}
\end{subfigure}
\begin{subfigure}{0.32\textwidth}
\centering
\includegraphics[width = \textwidth]{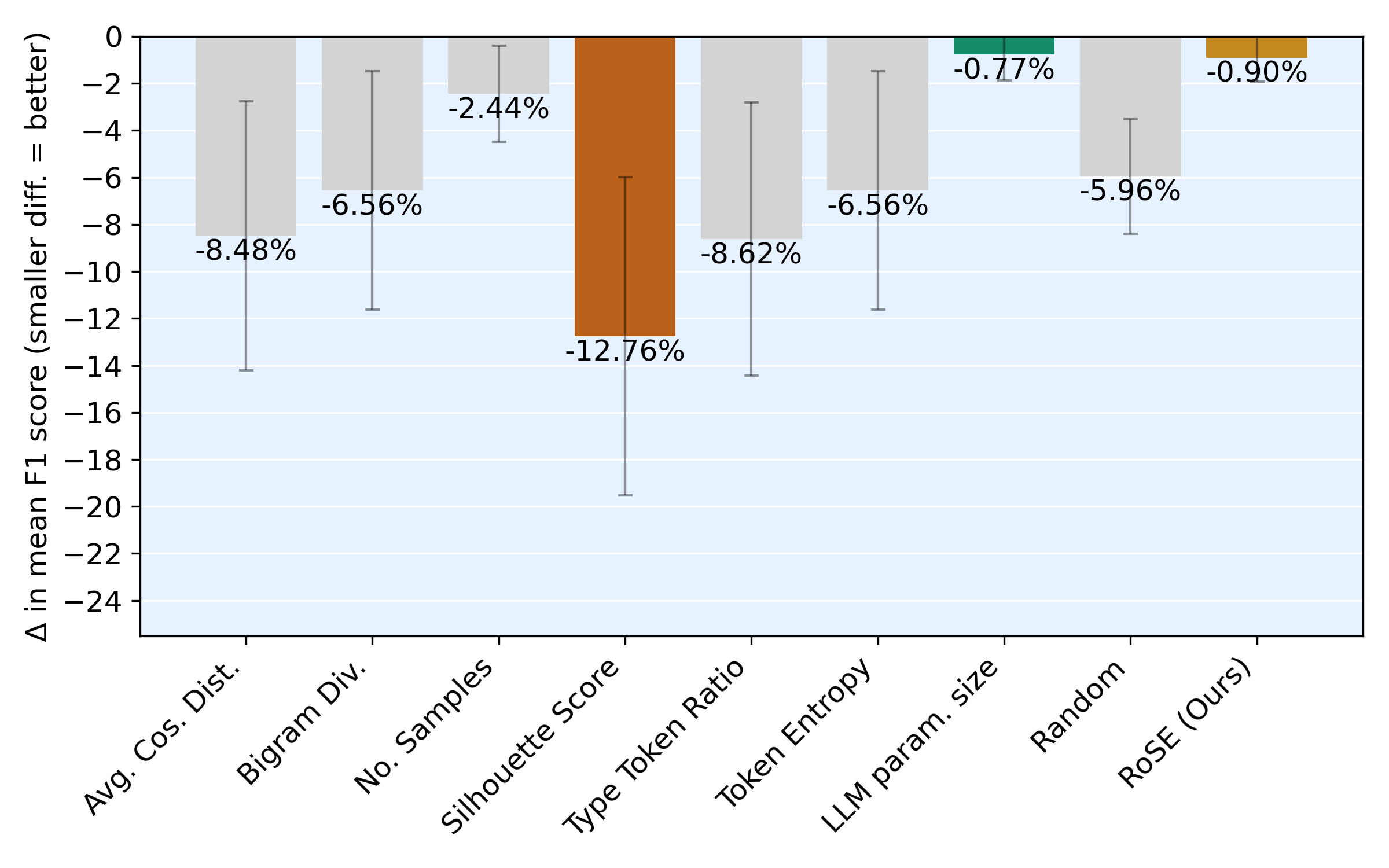}
\caption{Welsh.}
\label{fig:per_lang_cy}
\end{subfigure}
\begin{subfigure}{0.32\textwidth}
\centering
\includegraphics[width = \textwidth]{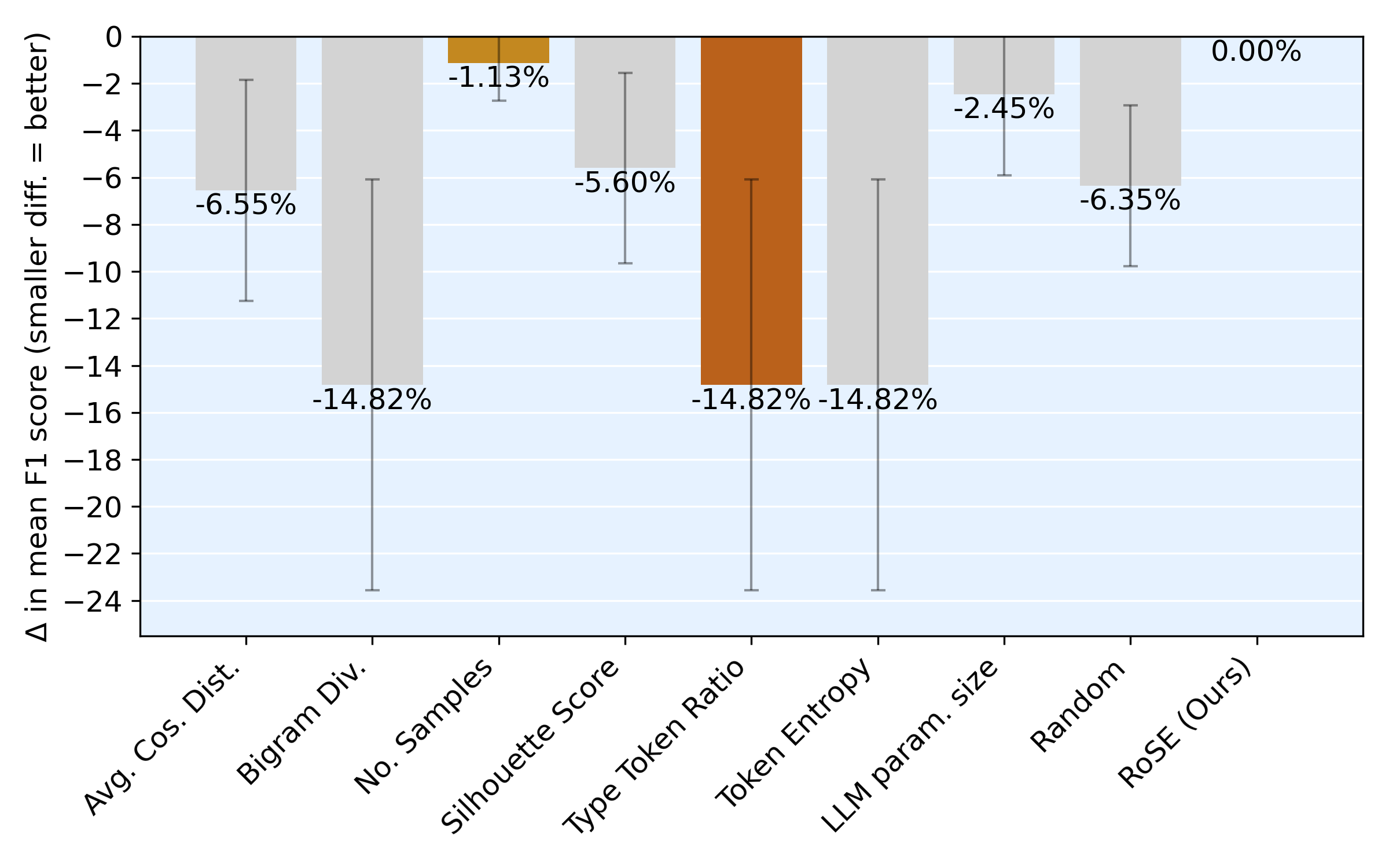}
\caption{Hebrew.}
\label{fig:per_lang_he}
\end{subfigure}
\begin{subfigure}{0.32\textwidth}
\centering
\includegraphics[width = \textwidth]{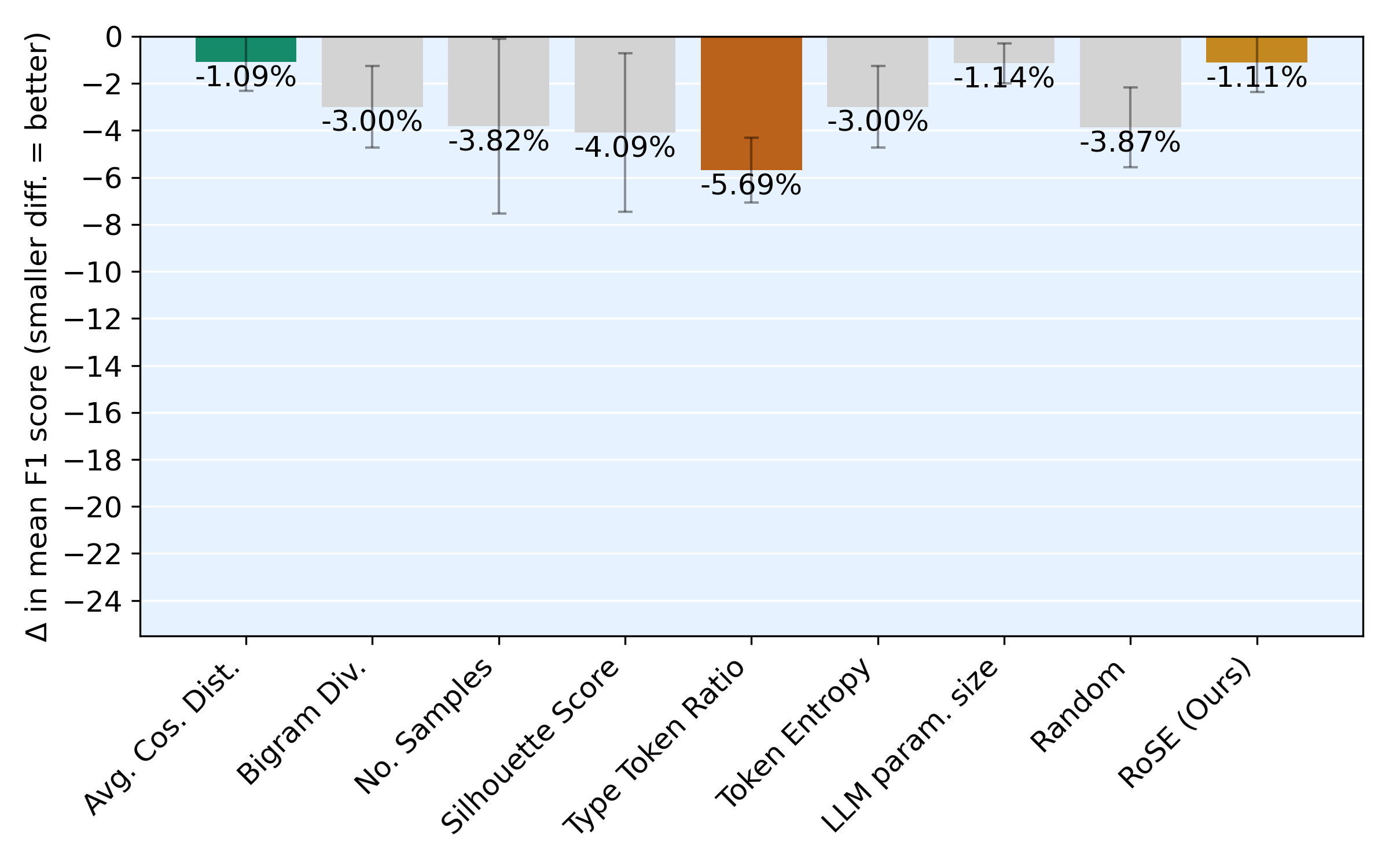}
\caption{Indonesian.}
\label{fig:per_lang_id}
\end{subfigure}
\begin{subfigure}{0.32\textwidth}
\centering
\includegraphics[width = \textwidth]{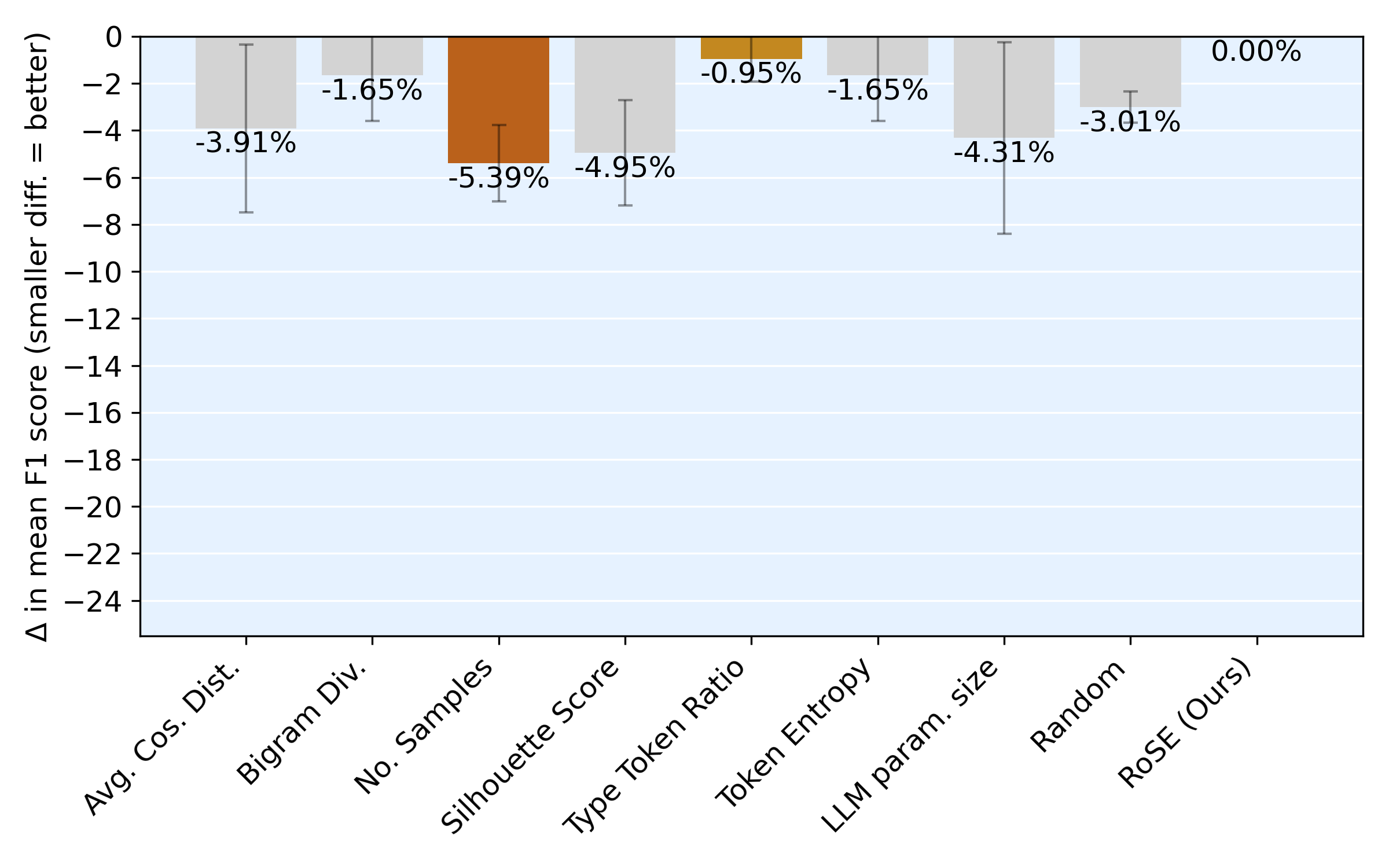}
\caption{German.}
\label{fig:per_lang_de}
\end{subfigure}
\begin{subfigure}{0.32\textwidth}
\centering
\includegraphics[width = \textwidth]{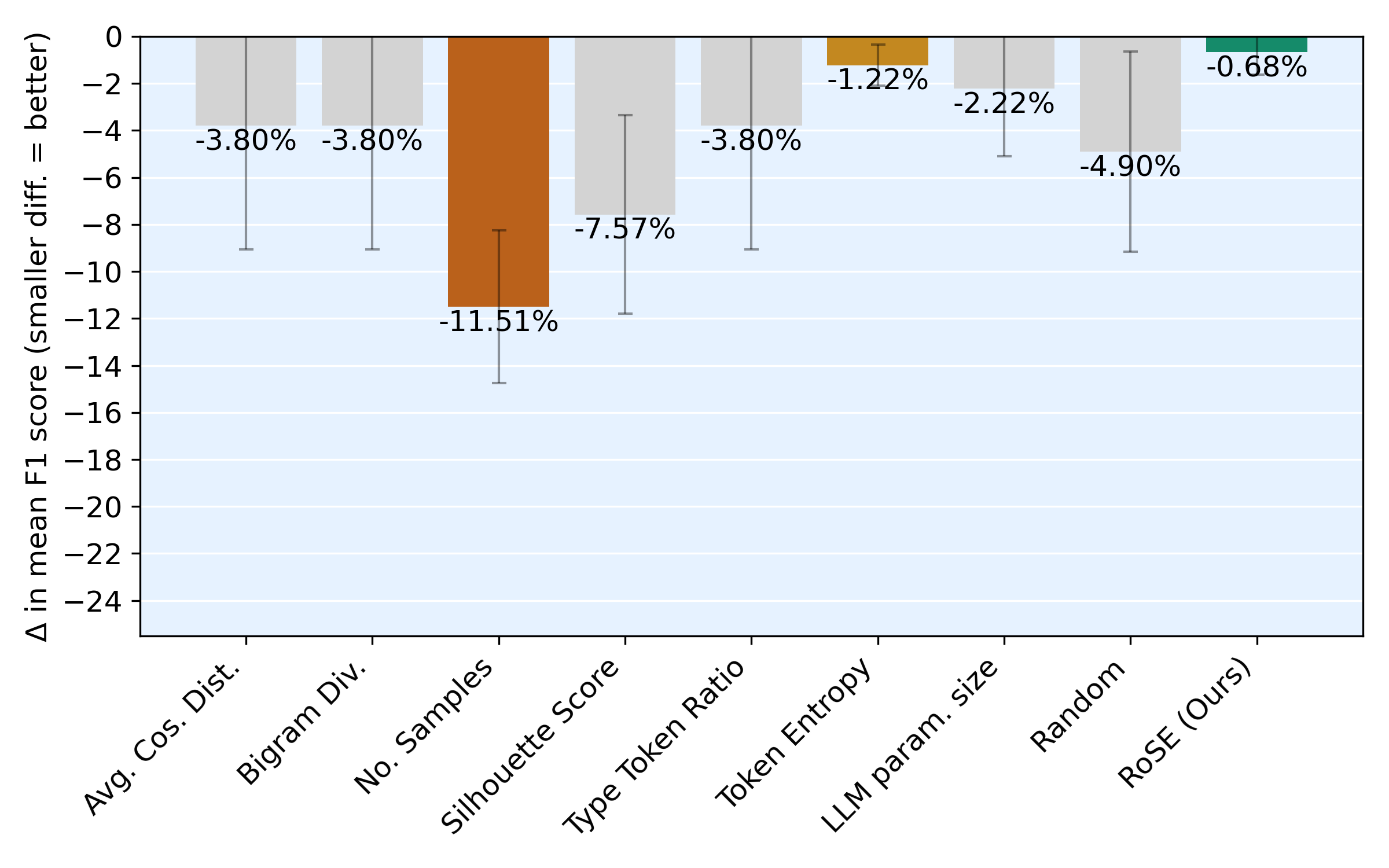}
\caption{English.}
\label{fig:per_lang_en}
\end{subfigure}
\begin{subfigure}{0.32\textwidth}
\centering
\includegraphics[width = \textwidth]{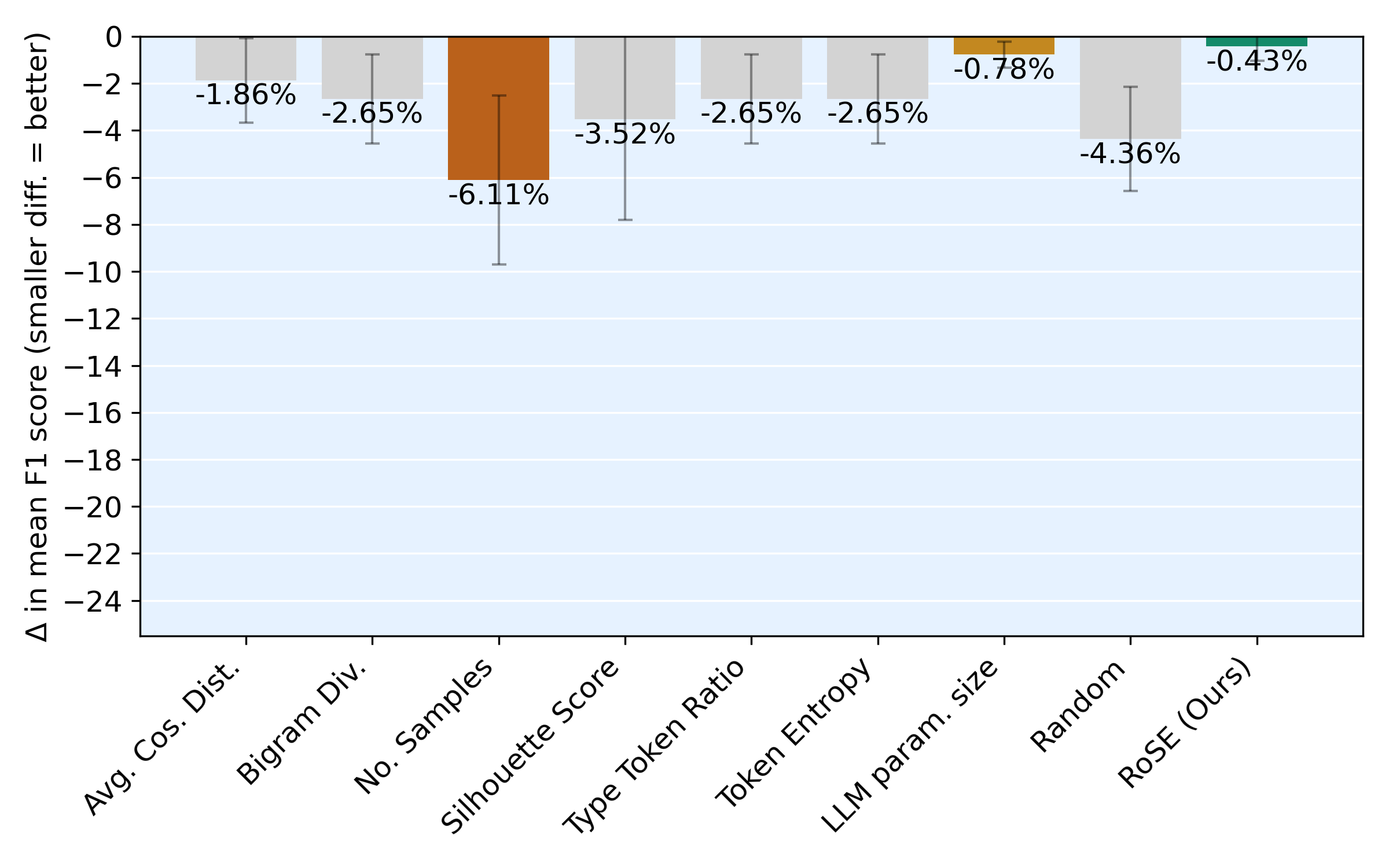}
\caption{Romanian.}
\label{fig:per_lang_ro}
\end{subfigure}
\begin{subfigure}{0.32\textwidth}
\centering
\includegraphics[width = \textwidth]{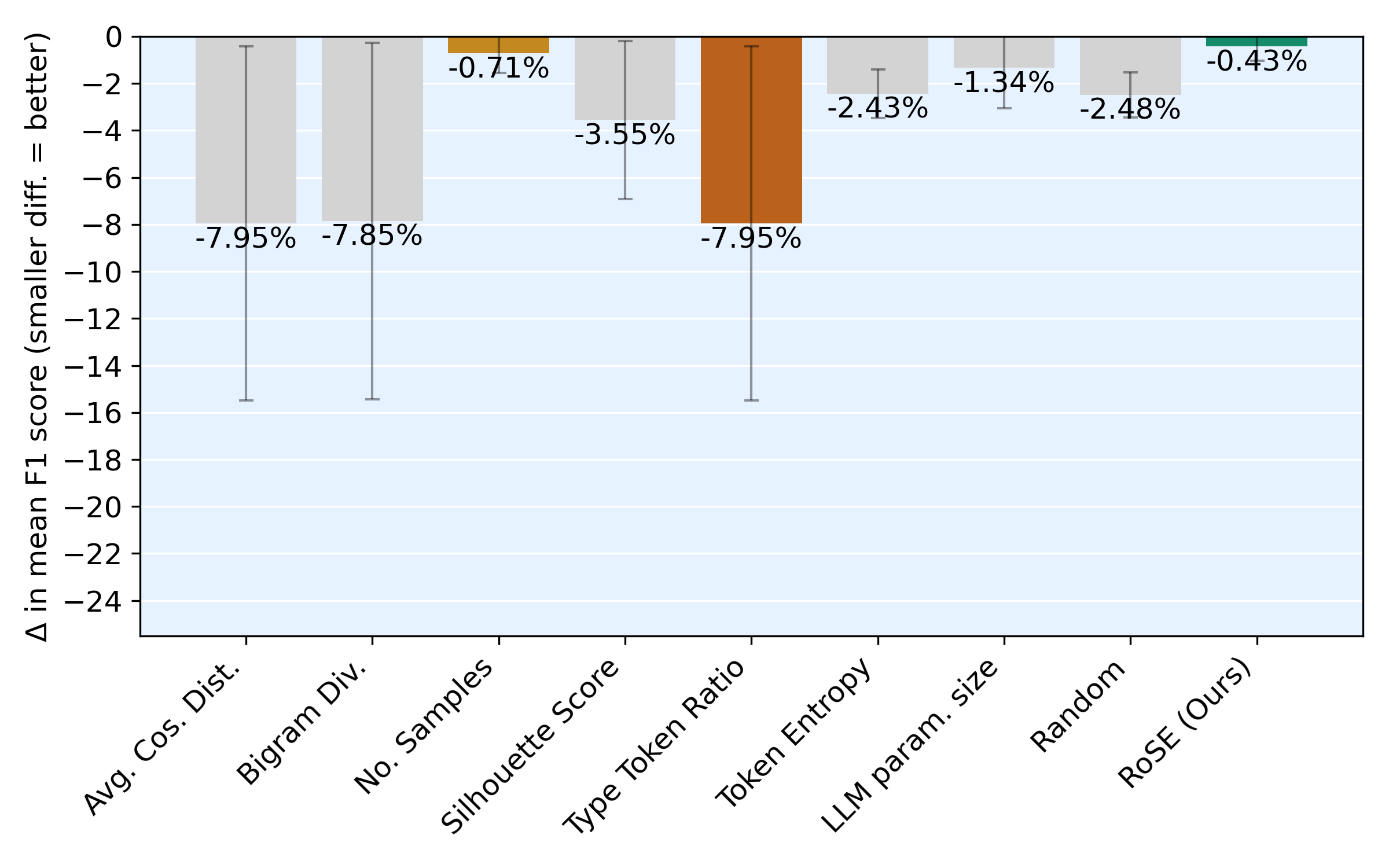}
\caption{Slovenian.}
\label{fig:per_lang_sl}
\end{subfigure}
\begin{subfigure}{0.32\textwidth}
\centering
\includegraphics[width = \textwidth]{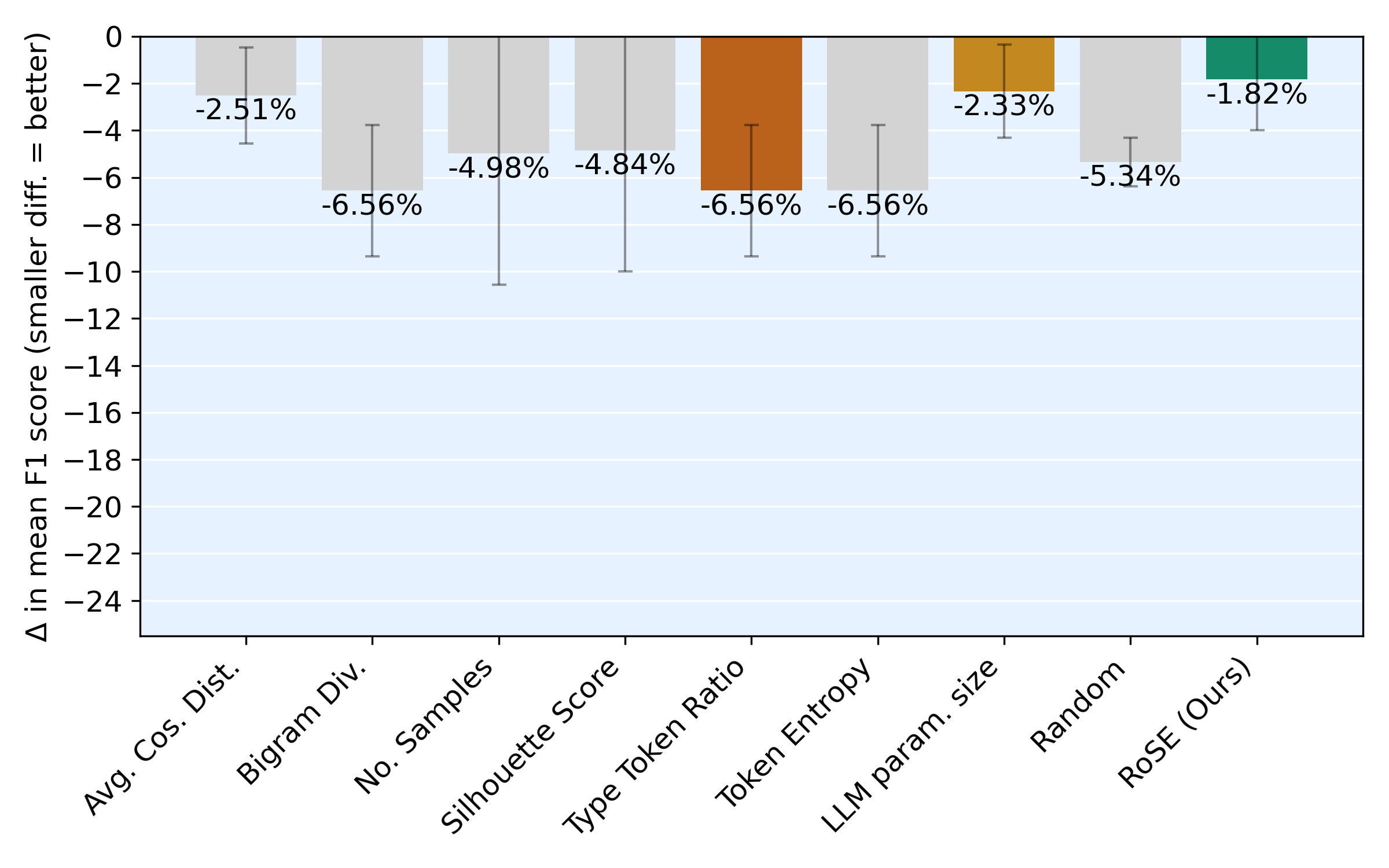}
\caption{Swahili.}
\label{fig:per_lang_sw}
\end{subfigure}
\begin{subfigure}{0.32\textwidth}
\centering
\includegraphics[width = \textwidth]{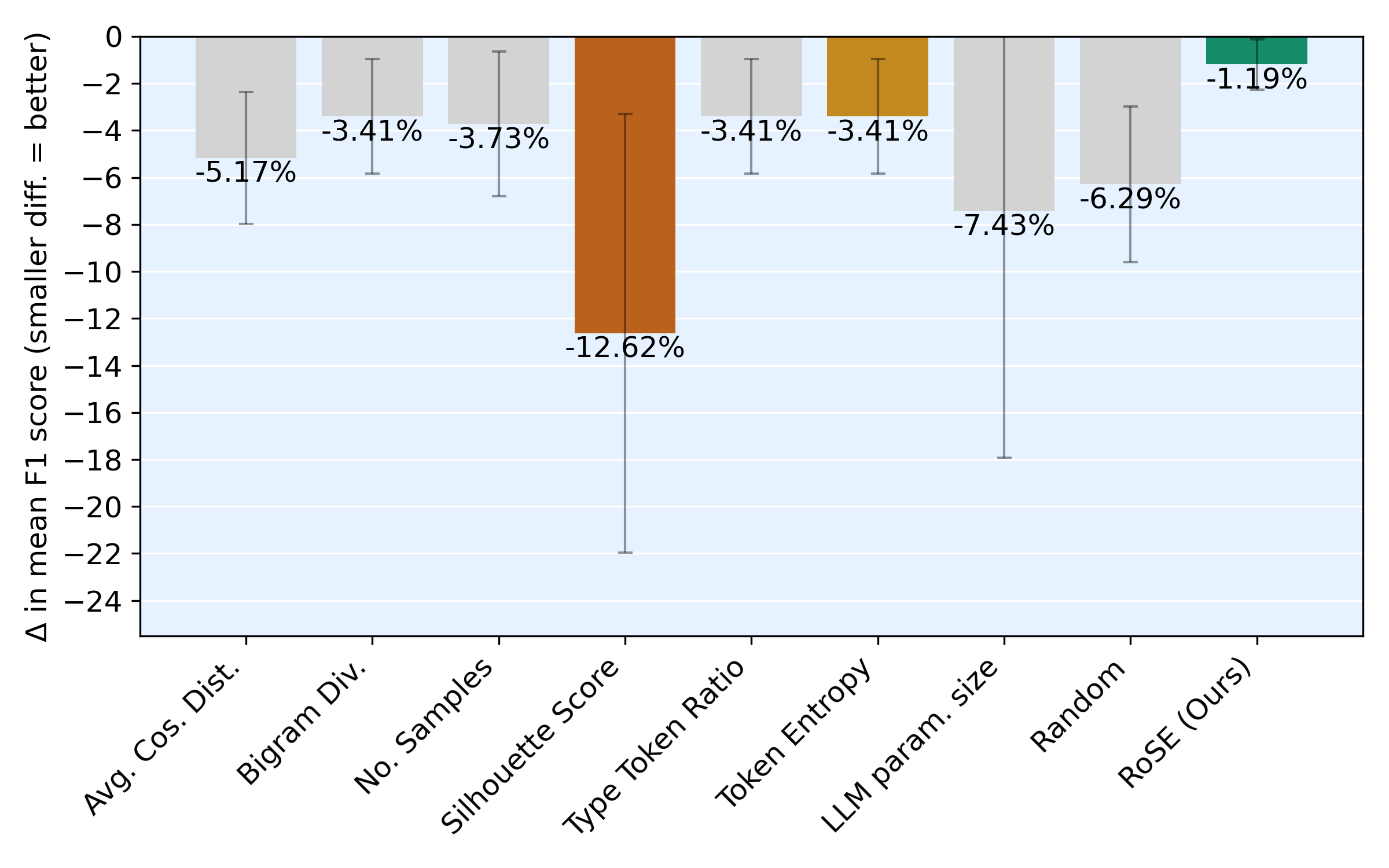}
\caption{Telugu.}
\label{fig:per_lang_te}
\end{subfigure}
\begin{subfigure}{0.32\textwidth}
\centering
\includegraphics[width = \textwidth]{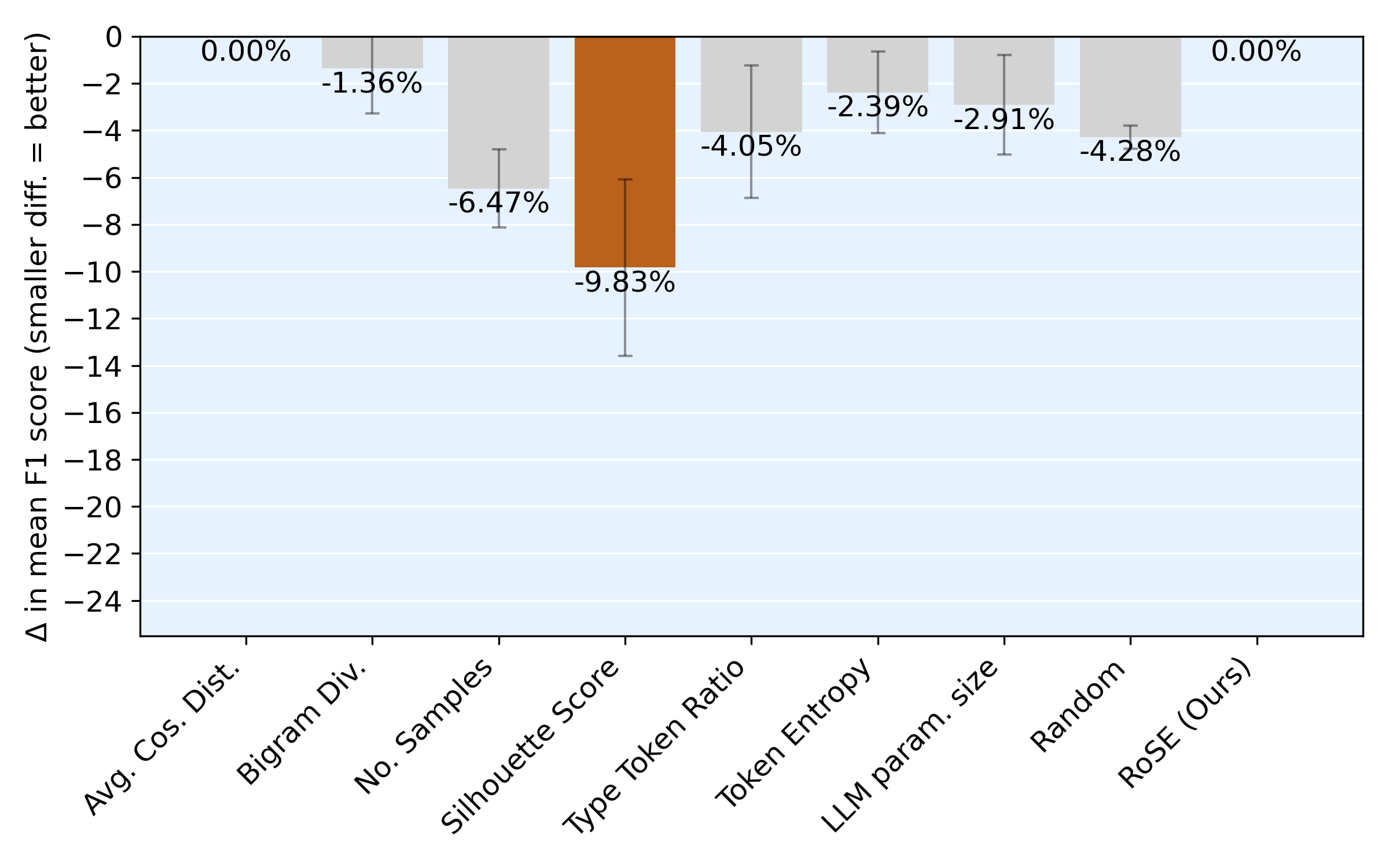}
\caption{Thai.}
\label{fig:per_lang_th}
\end{subfigure}
\caption{Comparison of proxy metrics for selecting the best LLM generator per language. Bars show the average gap in mean F1 score for models trained on the best generator selected by metrics vs. the optimal generator (smaller is better). The best metric is green, the second best is orange, and the worst is red. RoSE is the best proxy metric in 9 of 11 languages, and 2nd best for the other 2 cases. RoSE performs well across all languages regardless of the amount of resources per language.}
\label{fig:perf_per_lang}
\end{figure*}

We provide an evaluation of proxy metrics, excluding Llama 3 70B, in Figure~\ref{fig:results_main_ablt}. We provide additional visualisation for proxy metric performance per task in Figure~\ref{fig:combined-per-task} and per language in Figure~\ref{fig:perf_per_lang}. We provide the mean F1 scores on human test data for finetuned XLM-R per language-task in Tables~\ref{tab:human_f1_table_sentiment},~\ref{tab:human_f1_table_topic} and~\ref{tab:human_f1_table_intent}. We provide an overview of the performance and costs of each proxy metric used in Table~\ref{tab:proxy_metrics_summary}.

\end{document}